\documentclass[mnsc]{informs3b} 

\DoubleSpacedXI

\usepackage{natbib}
 \bibpunct[, ]{(}{)}{,}{a}{}{,}%
 \def\bibfont{\small}%
\TheoremsNumberedThrough     
\ECRepeatTheorems

\EquationsNumberedThrough    

\usepackage{subcaption}
\usepackage[bottom]{footmisc}
\usepackage{afterpage}
\usepackage{float}
\usepackage{xcolor}
\usepackage[bookmarks, hidelinks]{hyperref}
\hypersetup{
	bookmarksnumbered=true,     
	bookmarksopen=true,
	bookmarksopenlevel=1,    
	colorlinks=true,
	linkcolor={red!50!black},
	citecolor={blue!50!black},
	urlcolor={blue!80!black},
}
\usepackage{tabularx}
\usepackage{algorithm}  
\usepackage{algorithmic}
\usepackage{multirow}
\usepackage{graphicx}

\usepackage{enumitem}
\usepackage{caption,subcaption}
\usepackage{cleveref}

\RequirePackage{amsfonts}
\newcommand{\real}{\mathbb{R}}

\newcommand{\x}{\mathbf x}
\newcommand{\z}{\mathbf{z}}

\renewcommand{\h}{\mathbf{h}}
\newcommand{\g}{\mathbf{g}}

\newcommand{\s}{\mathcal{S}}

\newcommand{\D}{\mathcal{D}}
\newcommand{\M}{\mathcal{M}}

\newcommand{\loss}{\mathcal{L}}

\newcommand{\bd}{\boldsymbol}

\begin{document}

\newcolumntype{L}[1]{>{\raggedright\arraybackslash}p{#1}}
\newcolumntype{C}[1]{>{\centering\arraybackslash}p{#1}}
\newcolumntype{R}[1]{>{\raggedleft\arraybackslash}p{#1}}



\RUNTITLE{Privacy-Preserving Credit Risk Prediction with Alternative Data}

\TITLE{Privacy-Preserving Credit Risk Prediction with Alternative Data}

\ARTICLEAUTHORS{%
\AFF{}
\AUTHOR{Hongzhe Zhang}
\AFF{School of Management and Economics, The Chinese University of Hong Kong, Shenzhen (CUHK-Shenzhen), China} 
\AUTHOR{Jiarong Xu}
\AFF{School of Management, Fudan University, Shanghai, China} 
\AUTHOR{Jing He}
\AFF{Lerner College of Business and Economics, University of Delaware, Newark, Delaware}
\AUTHOR{Xiao Fang\thanks{Corresponding Author: Xiao Fang, \url{xfang@udel.edu}}}
\AFF{Lerner College of Business and Economics, University of Delaware, Newark, Delaware}
}

\ABSTRACT{Credit risk prediction, or how to develop predictive models that aid financial institutions in granting and managing consumer credit, is a critical problem in the consumer credit industry. Traditionally, financial institutions construct credit risk prediction models using borrowers' demographic, financial, and credit history data, collectively referred to as traditional data. Recent studies have demonstrated that alternative data, such as borrowers' mobile phone communication data, enable lenders to acquire fuller and more accurate profiles of borrowers' creditworthiness, thereby improving credit risk prediction performance. Nevertheless, alternative data are held by external entities independent of financial institutions. Directly sharing alternative data with financial institutions infringes on consumer privacy, yet existing credit risk prediction studies largely overlook this issue. To address this research gap, we define a new problem, namely privacy-preserving credit risk prediction with alternative data, which simultaneously considers three practical constraints for building credit risk prediction models with alternative data: the privacy-preserving constraint that protects consumer privacy, the model-confidentiality constraint that learns and stores the model centrally at the financial institution, and the lossless constraint that maintains the performance of the learned model. To solve this problem, we develop PrivacyCredit, a novel privacy-preserving machine learning method. We then theoretically demonstrate the privacy-preserving, model-confidential, and lossless properties of PrivacyCredit. Through extensive experiments using a real-world credit dataset linked with alternative data, we demonstrate the predictive value of securely incorporating alternative data into credit risk prediction and show that PrivacyCredit achieves the same predictive performance as the model learned from the insecure plaintext combination of traditional and alternative data. We further evaluate its model-confidentiality property and computational efficiency, showing that PrivacyCredit can be practically implemented for credit risk prediction.

}%



\def\HOOKtop{\vspace*{-25pt}}

\maketitle

%


\section{Introduction} \label{sec:intro}

Consumer credit, which refers to money loaned by financial institutions to consumers, is central to modern consumer finance services \citep{hurley2016credit}. The consumer credit market has continued to expand in recent years. Taking the U.S. as an example, credit card balances alone increased by \$44 billion to \$1.28 trillion in the fourth quarter of 2025, up 5.5\% from a year earlier, while other balances, including retail cards and consumer finance loans, increased by \$14 billion to \$564 billion.\footnote{See \href{https://www.newyorkfed.org/medialibrary/interactives/householdcredit/data/pdf/HHDC_2025Q4}{\url{https://www.newyorkfed.org/medialibrary/interactives/householdcredit/data/pdf/HHDC_2025Q4}} (last accessed on May 30, 2026).} Given the sheer size and rapid growth of the consumer credit market, credit risk prediction, i.e., how to predict the default probability of a consumer credit, has become a crucial problem for the consumer credit industry and attracted significant attention from academic researchers \citep{iyer2016screening,zhang2025latent}. An effective credit risk prediction method benefits all participants in the consumer credit market: it helps financial institutions identify risky credit applicants and reduce potential losses, while enabling creditworthy consumers to gain access to much-needed credit. 

Credit risk prediction is commonly formulated as a classification problem. Prior studies have applied a variety of classification methods to this problem, such as logistic regression, multilayer perceptron, support vector machine, and eXtreme Gradient-tree Boosting (XGBoost)  \citep{sinha2004evaluating, chen2016xgboost, thomas2017credit}. Among these methods, XGBoost has been widely used for credit risk prediction due to its interpretability and excellent predictive performance \citep{zhang2025latent}. Yet the effectiveness of credit risk prediction models depends not only on their underlying classification methods but also on the informativeness of the training data. Therefore, beyond improving classification algorithms, another way to enhance the performance of credit risk prediction is to enrich the training data with additional data from other sources.

Traditionally, financial institutions construct credit risk prediction models using borrowers' demographic, financial, and credit history data \citep{thomas2017credit}. These data are held by financial institutions and referred to as traditional data \citep{lu2023profit}.  Beyond traditional data, the widespread adoption of digital technologies has enabled the collection of various forms of alternative data that are informative for credit risk prediction, such as mobile phone communication data and e‑commerce transaction data. Alternative data allow financial institutions to develop more comprehensive and accurate profiles of borrowers’ creditworthiness.  As a result, both industry practices and academic studies have shown that the performance of credit risk prediction can be significantly improved by combining traditional data with alternative data \citep{berg2020rise, lu2023profit,lee2026benefits}.
For example, \citet{lu2023profit} demonstrate the predictive power of social media and mobile phone communication data for credit risk assessment, whereas \citet{berg2020rise} and \citet{lee2026benefits} empirically show that borrowers’ digital footprints and retail transaction records are predictive of consumers’ credit card repayment behavior.

Alternative data are stored at external entities independent of financial institutions, such as mobile service providers and e‑commerce platforms. Directly sharing such data with financial institutions for credit risk prediction severely infringes on consumer privacy \citep{world2018data}. One class of solutions seeks to address this challenge by modifying alternative data before sharing them with financial institutions so that certain privacy‑preserving objectives are satisfied. This line of work is generally referred to as privacy‑preserving data publishing (PPDP) \citep{fung2010privacy}. 
For example, the $k$-anonymity method modifies original data such that each record has at least $k-1$ other records with identical attribute values \citep{sweeney2002k}, thereby preventing individual records from being uniquely identified. As another example, $\epsilon$-differential privacy methods add random noise to original data, where parameter $\epsilon$ controls the amount of allowed information leakage \citep{dwork2025differential}. 
Nevertheless, PPDP methods suffer from two key limitations. First, they can only prevent privacy leakage to a limited extent and  remain vulnerable to privacy breaches \citep{cheng2021secureboost}. For instance, $\epsilon$-differential privacy must allow a certain level of information leakage to ensure that useful models can still be learned from the modified data \citep{wood2018differential}. Second, modifying data inevitably reduces data utility, which in turn degrades the predictive performance of models (e.g., credit risk prediction models) trained on the modified data \citep{li2009tradeoff}.\footnote{Recent work attempts to preserve data utility when applying PPDP methods; however, these approaches typically maintain only the statistical properties of the dataset and fail to guarantee the predictive performance of machine learning models trained on the modified data (e.g., \citealp{bi2023distribution,qu2026shapley}).}  

Federated learning has emerged as a promising paradigm that facilitates collaborative machine learning among multiple parties while preserving each participant's data privacy \citep{kairouz2021advances, bi2024understanding}. In federated learning, multiple parties (e.g., a bank and a mobile service provider) jointly train a machine learning model while keeping their data locally stored and processed \citep{li2020federated}. In contrast to PPDP methods, federated learning obviates the need for direct data exchange among participating parties, thereby offering stronger protection of data privacy. However, existing federated learning methods train models in a distributed manner across participating parties, implying that each party inadvertently gains access to part or even all of the trained model \citep{wu20privacy,cheng2021secureboost}. This feature creates a critical challenge for financial institutions seeking to use federated learning to build credit risk prediction models. In fact, financial institutions treat their credit risk prediction models as closely guarded trade secrets and therefore require the learned model to remain strictly confidential \citep{hurley2016credit}. If such models were disclosed to external parties, malicious borrowers could potentially reverse engineer the decision boundaries and strategically manipulate their observable attributes to obtain credit approval despite high underlying risk. Such strategic manipulation would undermine the effectiveness of credit risk screening and increase financial institutions’ exposure to credit losses \citep{ghalme2021strategic}. As a result, the decentralized model ownership inherent in existing federated learning methods conflicts with the requirement that credit risk prediction models must be centrally learned and stored at financial institutions, rendering these methods inapplicable to securely leveraging alternative data for credit risk prediction.

Taken together, although combining traditional and alternative data can substantially enhance the performance of credit risk prediction, insecure use of alternative data severely compromises consumer privacy. As discussed above, existing privacy-preserving methods have limitations in addressing this challenge.
In response, we define a new problem, namely Privacy-preserving Credit Risk Prediction with Alternative data (PCRPA), which simultaneously considers three practical constraints when building a credit risk prediction model with alternative data: the privacy-preserving constraint that protects consumer privacy, the model-confidentiality constraint that learns and stores the model centrally at the financial institution, and the lossless constraint that preserves the predictive performance. We then propose PrivacyCredit, a novel privacy-preserving machine learning method, to solve this problem.

To protect consumer privacy, PrivacyCredit keeps each party's data local and permits the exchange of only encrypted or randomly masked data among parties. The primary technical challenge in realizing this design is that the financial institution must train a credit risk prediction model and perform credit risk inference using encrypted data. However, model training and inference are difficult to execute directly in an encrypted environment, particularly for sophisticated machine learning–based credit risk prediction models. The key methodological innovation of this study addresses this challenge by enabling the financial institution to train and use an effective credit risk prediction model entirely in an encrypted environment while satisfying all three constraints of the PCRPA problem.
Unlike PPDP methods, which offer only limited privacy protection and degrade model performance, PrivacyCredit uses encryption and random masking to provide complete data privacy protection without sacrificing model performance. Different from existing federated learning methods, where the learned model is distributed across participating parties, PrivacyCredit learns and stores the credit risk prediction model centrally at the financial institution, thereby ensuring model confidentiality.

Our study contributes to the literature by proposing a new problem of privacy-preserving credit risk prediction with alternative data and developing a novel method to solve this problem. We also theoretically establish the privacy-preserving, model-confidential, and lossless properties of the proposed method. Through extensive experiments using a real-world credit dataset linked with alternative data, we demonstrate the predictive value of alternative data for credit risk prediction and show that the model trained by our method achieves the same predictive performance as the model trained on the insecure plaintext combination of traditional and alternative data. We further evaluate the model-confidentiality and computational efficiency of our method, showing that it can be practically implemented for real-world credit risk prediction.

 \section{Related Work}
\label{sec:rw}

Two research streams are related to our study: existing credit risk prediction models with alternative data and privacy-preserving machine learning methods over vertically partitioned data. In the following subsections, we review representative methods from each stream and discuss the key novelties of our study.

\subsection{Credit Risk Prediction with Alternative Data}

Prior research typically formulates credit risk prediction as a classification problem and addresses it using classification methods \citep{thomas2005survey, lessmann2015benchmarking, zhang2025latent}. In this formulation, each borrower is represented by a set of variables that characterize their creditworthiness and labeled according to their observed repayment outcome. Traditionally, financial institutions construct these variables from borrowers' credit reports, including demographic, financial, and credit history data, which are collectively referred to as traditional data. Recent research has shown that integrating training data with alternative data enables lenders to develop more comprehensive and accurate borrower profiles, thereby significantly enhancing credit risk prediction performance \citep{berg2020rise, lu2023profit, lee2026benefits}.

One important type of alternative data is mobile phone usage data. Prior studies have demonstrated that an individual's communication patterns, mobility traces, and other phone usage data can be used to infer their socioeconomic status \citep{blumenstock2015predicting}. Building on this insight, \citet{ma2018new} show that mobility and app-usage patterns are significantly correlated with loan default status and can be utilized for loan default prediction. \citet{yang2019understanding} analyze more than 1.5 billion calls among over 11 million online lending users. They document systematic differences between default and non-default borrowers in communication behaviors, including the number of contacts and intensity of calls during the loan application period. \citet{chen2025digital} demonstrate that mobile data are informative about borrowers' financial well-being, making them useful for loan default prediction. Specifically, they show that financial information contained in text messages, together with borrowers' location histories, can provide objective proxies for their financial conditions.

Recent studies also show strong evidence that grocery and retail activities contain informative signals for evaluating borrowers' creditworthiness. For example, \citet{lee2025using} demonstrate that grocery items purchased by a borrower are predictive of their credit card repayment behavior. In particular, purchases of vice products (e.g., cigarettes) and convenience foods (e.g., mortadella) are strongly associated with a higher likelihood of delinquency, while purchases of healthier food and ingredients for home cooking (e.g., fresh beans) are linked to lower delinquency risk. \citet{lee2026benefits} extend this line of research by examining a broader set of retail footprints beyond purchased items, including transaction timestamps, listed and paid prices, and transaction channels. Their study reproduces the findings of \citet{lee2025using} using a newly collected large-scale real-world dataset, thereby providing evidence of external validity across diverse economic and cultural contexts.

Other forms of alternative data have also been explored for credit risk prediction. For example, \citet{berg2020rise} investigate borrowers' traceable digital activities on websites and digital services (e.g., device type and operating system), commonly referred to as digital footprints. They show that these digital footprints serve as proxies for borrowers' financial characteristics and can significantly improve the performance of credit risk prediction models. In addition, \citet{lu2023profit} find that borrowers' social media presence and the sentiment of their microblog posts are valuable for credit risk prediction. \citet{djeundje2021enhancing} examine psychometric variables, such as conscientiousness, emotional stability, openness to experience, and integrity, and demonstrate that these traits can improve the predictive accuracy of credit scoring systems and help reduce financial exclusion.

Although alternative data are valuable for credit risk prediction, using them insecurely seriously infringes on consumer privacy \citep{world2018data}. This raises a central challenge: how to effectively leverage alternative data for credit risk prediction while preserving consumer privacy. Recent developments in privacy-preserving machine learning methods for vertically partitioned data offer potential solutions to this challenge, which we review next.

\subsection{Privacy-Preserving Machine Learning for Vertically Partitioned Data}

Depending on whether data are partitioned across organizations in the entity space or the feature space, two primary data partitioning scenarios arise: horizontally partitioned data and vertically partitioned data \citep{aggarwal2008general}. In horizontally partitioned data, different organizations possess the same set of features for different entities \citep{kantarcioglu2004privacy,mcmahan2017communication,so2020scalable}. In vertically partitioned data, different organizations own different sets of features for the same entities.
Since the financial institution and the alternative data owner hold different feature sets for the same users, our PCRPA problem falls into the scenario of vertically partitioned data. Therefore, we focus on reviewing privacy-preserving machine learning techniques designed for vertically partitioned data.

\subsubsection{Privacy-Preserving Data Publishing.}
A straightforward approach is to sanitize datasets before transferring them, which falls within the line of Privacy-preserving Data Publishing (PPDP, \citealp{fung2010privacy}). One of the most well-known PPDP models is $k$-anonymity, proposed by \citet{sweeney2002k}. By applying generalization and suppression to quasi-identifiers (QIDs), i.e., non-sensitive attributes that do not directly identify individuals (e.g., age, sex, and ZIP code), the $k$-anonymity model ensures that each record in the released dataset is indistinguishable from at least $k-1$ other records with respect to the QIDs. Unlike $k$-anonymity, which only modifies quasi-identifiers, $l$-diversity requires each anonymized group to contain at least $l$ ``well-represented'' sensitive values \citep{machanavajjhala2007diversity}. \citet{li2007t} further propose the $t$-closeness model, which constrains the distribution of sensitive attributes within each group to be close to that of the entire dataset. In addition, other anonymization approaches have been developed to defend against specific types of attacks, such as classification attacks \citep{li2009against}, regression attacks \citep{li2014digression}, and composition attacks \citep{li2016hybrid}. 

Nevertheless, such anonymization techniques lack a rigorous theoretical foundation and can only defend against known attacks \citep{chen2021privacy, mohammed2011differentially}. To address this limitation, \citet{dwork2025differential} proposes the $\epsilon$-differential privacy model, which provides a rigorous and provable privacy guarantee for protected data. $\epsilon$-differential privacy injects random noise into original data, where the privacy loss parameter $\epsilon$ controls the amount of allowed information leakage. This differential privacy model has been further incorporated into PPDP methods to provide stronger privacy guarantees. For example, \citet{mohammed2011differentially} introduce a generalization-based algorithm to release a noisy contingency table, with the injected noise satisfying the differential privacy requirement. \citet{soria2014enhancing} further propose first applying the $k$-anonymity method to the original dataset before injecting noise, thereby reducing the amount of noise required to achieve $\epsilon$-differential privacy and improving the utility of the modified dataset. Despite these advances, modifying data through PPDP methods inevitably reduces data utility, and models trained on the modified data therefore suffer from degraded predictive performance \citep{li2009tradeoff}. Moreover, PPDP methods can only prevent information leakage to a limited extent \citep{cheng2021secureboost}. For instance, many anonymization techniques are designed to defend against specific attacks and remain vulnerable to previously unseen attack strategies \citep{mohammed2011differentially}. Even differential privacy must allow a certain degree of information leakage in order to preserve the utility of the released data \citep{wood2018differential}.

Besides modifying existing data as discussed above, another line of PPDP research focuses on synthetic data generation, also referred to as synthetic data publishing. Exemplar studies include \citet{zhang2017privbayes} and \citet{ho2021dp}, which construct statistical or generative models from the original dataset and release sampled synthetic records, using Bayesian networks and generative adversarial networks, respectively. However, this line of PPDP techniques does not preserve one-to-one correspondence with the original records, making it difficult to link a specific synthetic record to other datasets or model outputs \citep{el2020evaluating}. In our setting, by contrast, the alternative data must remain linkable to users' credit data, so synthetic-data-based PPDP is not applicable to our purpose.

\subsubsection{Vertical Federated Learning.}
Federated learning represents another stream of research that enables multiple parties to collaboratively build machine learning models while preserving user privacy and data confidentiality at each party \citep{yang2019federated,bi2024understanding}.
Existing federated learning research predominantly focuses on horizontally partitioned settings. In contrast, vertical federated learning (VFL), which allows multiple participants holding different features of the same users to jointly train a prediction model without revealing their raw data, has received relatively less attention. Recently, several studies have begun to address this challenge by developing VFL frameworks.
For example, \citet{hardy2017private} propose a VFL framework that enables two parties to collaboratively train a logistic regression model using an additively homomorphic encryption scheme. Building on this work, \citet{yang2019quasi} adopt a quasi-Newton method to reduce the communication complexity of training the federated logistic regression model, and \citet{yang2019parallel} further remove the third‑party coordinator required in earlier frameworks. Beyond logistic regression, other machine learning models have also been incorporated into VFL frameworks. For instance, \citet{cheng2021secureboost} develop SecureBoost, a tree‑boosting framework for vertical federated learning, and theoretically demonstrate that it achieves the same predictive accuracy as its non‑federated counterpart. \citet{gu2020federated} focus on nonlinear learning with kernels and propose a federated kernel method that achieves both high efficiency and scalability. In addition, \citet{wu20privacy} develop a privacy‑preserving decision tree framework for vertical federated learning that avoids disclosing intermediate information, thereby providing stronger privacy guarantees.

Moreover, some studies focus on designing VFL methods for specific application scenarios. For instance, data scarcity has attracted considerable attention. \citet{kang2022fedcvt} propose a semi-supervised learning approach to improve VFL performance when the number of overlapping samples across participating parties is limited. \citet{zhang2024toward} address another form of data scarcity, limited labeled data, and introduce a pseudo-label-guided consistency mechanism to effectively exploit unlabeled samples. Another line of work focuses on computational efficiency. In this vein, \citet{zhang2021asysqn} develop an asynchronous VFL framework that allows participating parties to update the model asynchronously, thereby improving computational efficiency and better utilizing heterogeneous computing resources across parties. \citet{xi2025private} further investigate efficient sample alignment in multi-client settings and propose a lightweight delegated Private Set Intersection scheme for sample intersection among multiple clients.  Other studies examine fairness, incomplete data, and data valuation in VFL. \citet{qi2022fairvfl} propose a privacy-preserving approach that learns unified and fair sample representations from decentralized feature fields while maintaining data confidentiality. \citet{liao2025privacy} focus on missing features and propose a tensor-decomposition-based network that learns both intra- and inter-client feature information to improve VFL performance. \citet{han2026data} study data valuation in VFL and propose a method to evaluate each data party's contribution, thereby enabling task parties to identify and select the data parties that contribute most to the focal task.

However, regardless of the specific methods or application scenarios considered, existing VFL approaches share a common design feature: the model is trained and maintained in a distributed manner across participating parties. Such a design conflicts with an important practical requirement in building credit risk prediction models. In fact, financial institutions treat their credit risk prediction models as closely guarded trade secrets and therefore require the learned model to remain confidential. As a result, they may be reluctant to adopt existing VFL methods because these frameworks cannot ensure that the model is centrally learned and stored at the financial institution.

\subsection{Key Novelties of Our Study}

\begin{table}[]
\TABLE
{Comparison between Our Method and Existing Methods \label{tab:method_compare}}
{ \resizebox{\textwidth}{!}
{
\begin{tabular}{L{180pt}C{90pt}C{90pt}C{80pt}}
\hline
& \begin{tabular}[c]{@{}c@{}}Satisfying\\ Privacy-preserving \\ Constraint\end{tabular} & \begin{tabular}[c]{@{}c@{}}Satisfying\\ Model-confidential \\ Constraint\end{tabular} & \begin{tabular}[c]{@{}c@{}}Satisfying\\ Lossless\\ Constraint\end{tabular} \\ \hline
\up Insecure Combination of Traditional and Alternative Data, e.g., \citet{berg2020rise}
& No            & Yes            & Yes  \\
\up Privacy-Preserving Data Publishing Methods, e.g., \citet{soria2014enhancing} 
& No           & Yes            & No   \\
\up Vertical Federated  Learning Methods, e.g., \citet{cheng2021secureboost}
& Yes           & No             & Yes  \\
\up\down Our Method                                               & Yes           & Yes            & Yes  \\ 
\hline
\end{tabular}
}
}
		{}
\end{table}

Our literature review suggests the following research gaps, as summarized in Table~\ref{tab:method_compare}. First, recent studies have shown that effective credit risk prediction methods should leverage both traditional and alternative data; however, they largely overlook the fact that insecure use of alternative data seriously infringes on consumer privacy. To address this gap, we introduce a new problem, namely Privacy-preserving Credit Risk Prediction with Alternative data (PCRPA), which simultaneously considers three realistic constraints for building credit risk prediction models: the privacy-preserving constraint that protects consumer privacy, the model-confidentiality constraint that learns and stores the model centrally at the financial institution, and the lossless constraint that preserves the predictive performance.

Second, none of the existing privacy-preserving machine learning methods can address this problem while satisfying all three constraints simultaneously. PPDP methods provide only limited privacy protection and degrade the performance of the learned model, whereas federated learning methods violate the model-confidentiality constraint required by financial institutions. In response, we propose a novel privacy-preserving machine learning method, namely PrivacyCredit, to solve the PCRPA problem. PrivacyCredit enables financial institutions to learn a credit risk prediction model and infer users' credit risk using traditional and alternative data while preserving consumer privacy, maintaining the confidentiality of the learned model, and achieving the same predictive performance as the model learned from the insecure plaintext combination of traditional and alternative data. Together, the formulation of the new and important PCRPA problem and the development of PrivacyCredit as a solution to this problem, constitute the contributions of our study.

\section{Problem Formulation}
\label{sec:formulation}

Let $F$ represent a financial institution (e.g., a bank) and $A$ denote an owner of alternative data. 
For example, $A$ can be an e-commerce firm, who holds users' online shopping data, or a mobile service provider, who stores users' communication records. 
Let $V=\{v_1,v_2,\dots,v_n\}$ denote a set of $n$ common users between $F$ and $A$. Financial institution $F$ owns traditional data: $X^F\in \real^{n \times d^F}$ and $\bd{y}$. Each row $X^F_{i\cdot}$ of matrix $X^F$ represents user $v_i$'s $d^F$ traditional credit-related characteristics, such as annual income; $y_i\in \bd y$ denotes $v_i$'s credit status, where $y_i=0$ represents no delinquency (or no default) and $y_i=1$ means delinquency (or default), $i=1,2,\dots,n$. In addition, $A$ owns alternative data $X^A\in \real^{n \times d^A}$. Each row $X^A_{i\cdot}$ of matrix $X^A$ represents user $v_i$'s $d^A$ alternative features, such as the average number of calls in a month. There is no overlap between traditional characteristics in $X^F$ and alternative features in $X^A$. $F$ aims to learn a credit risk prediction model using both traditional data $(X^F, \bd{y})$ and alternative data $X^A$ in a privacy-preserving manner. Meanwhile, the credit risk prediction model must be learned and stored at $F$ and no information about the model can be leaked to any other party because financial institutions treat their credit risk prediction models as closely-guarded trade secrets \citep{hurley2016credit}. 
Now, we can formally define the Privacy-preserving Credit Risk Prediction with Alternative data (PCRPA) problem.
 
\begin{definition} [\bf PCRPA Problem]
 A financial institution $F$ privately owns traditional data ($X^F$, $\bd y$) and an alternative data owner $A$ privately holds alternative data $X^A$. $F$ aims to learn a credit risk prediction model $\M$ that must satisfy the following three constraints.
\\\indent(1) Privacy-preserving constraint: No traditional data at $F$ can be leaked to $A$ and no alternative data at $A$ can be leaked to $F$. 
\\\indent(2) Model-confidential constraint: The model $\M$ must be learned and stored at $F$ and no information of $\M$ can be leaked to any other party.
\\\indent(3) Lossless constraint: The performance of $\M$ should be the same as that of the credit risk prediction model $\M^{\circ}$ learned from the insecure combination of ($X^F$, $\bd y$) and $X^A$.\footnote{Insecure combination means that plaintext $X^A$ is transferred from $A$ to $F$ and then combined with ($X^F$, $\bd y$). Clearly, learning model $\M^{\circ}$ from the insecure combination of ($X^F$, $\bd y$) and $X^A$ violates the privacy-preserving constraint.}
\end{definition} 

To solve the PCRPA problem, we need to tackle the following methodological challenges. First, how to learn a credit risk prediction model from data distributed at $F$ and $A$, while satisfying the three constraints at the same time. Second, how to predict credit risks for new borrowers, whose data are distributed at $F$ and $A$, while maintaining the three constraints at the same time. In response to these challenges, we propose a novel privacy-preserving credit risk prediction method to solve the PCRPA problem. 

Because the financial institution and the alternative data owner possess different features for the same set of users, the PCRPA problem falls into the setting of vertically partitioned data. Under this setting, a privacy-preserving method is designed for a specific machine learning model, as the privacy-preserving protocol must be designed around the model’s learning and inference procedures \citep{yang2019federated_book}. For example, \citet{hardy2017private} develop a privacy-preserving method for learning and inference with logistic regression on vertically partitioned data, whereas \citet{liu2020federated} design a privacy-preserving method specifically for random forests. In this study, we propose a privacy-preserving method for XGBoost \citep{chen2016xgboost}, which is the most widely used and effective machine learning model for credit risk prediction \citep{li2020xgboost,zhang2025latent}. 
In the following, we first review XGBoost and a widely used encryption scheme for privacy-preserving machine learning and then propose the two building blocks of our method, each of which addresses one of the aforementioned methodological challenges. 

\section{Background}
\label{sec:background}

\subsection{Vanilla XGBoost}
\label{sec:background:xgboost}

Given a dataset of $n$ training instances $\D=\{(\x_i,y_i)|\x_i\in\real^d, y_i\in\{1,0\}, i=1,2,\dots,n\}$, where $\x_i$ is a vector of $d$ features of instance $i$ and $y_i$ denotes its class label, XGBoost learns $T$ regression trees from $\D$ \citep{chen2016xgboost}.\footnote{
	XGBoost takes numerical features as inputs, and categorical features can be converted to numerical features through one-hot encoding \citep{chen2016xgboost}.
} We denote the $t$-th regression tree as $\tau_{t}$,  $t=1,2,\dots,T$, and the weight of its $k$-th leaf as $\omega_{t,k}$, where $k=1,2,\dots,K$ and $K$ is the number of its leaves. Tree $\tau_{t}:\real^d\rightarrow\real$ takes an instance's features as inputs and returns the weight of the leaf that the instance is classified into. That is,   
\begin{equation}
	\label{eq:tau_t}
	\tau_{t}(\x_i)=\omega_{t,q_t(\x_i)},
\end{equation}
where function $q_t:\real^d\rightarrow\{1,2,\dots,K\}$ classifies the instance into one of the $K$ leaves of $\tau_{t}$. For instance $i$, XGBoost predicts its probability of belonging to class 1, i.e., $y_i=1$, by
\begin{equation}
	\label{eq:xgboost}
	{\theta}_i=\sum_{t=1}^{T}\tau_{t}(\x_i), \quad p_i=\sigma({\theta}_i),
\end{equation}
where ${p}_i$ and ${\theta}_i$ denote the estimated probability and log-odds for $y_i=1$, respectively, $\sigma(\cdot)$ is the logistic function, and $\sigma(x)=\frac{1}{1+e^{-x}}$. 

XGBoost learns from training data to construct one tree at each iteration. Specifically, tree $\tau_{t}$ at iteration $t$ is learned by minimizing the following loss:
\begin{equation}
	\label{eq:xgboost_loss}
	\loss^{(t)}=\sum_{i}[l(y_i,\theta_i^{(t-1)})+g_{t,i}\tau_t(\x_i)+\frac{1}{2}h_{t,i}\tau^2_t(\x_i)]+\Omega(\tau_t),
\end{equation}
where regularization term $\Omega(\tau_t)=\gamma K+\frac{1}{2}\lambda\sum_k\omega_{t,k}^2$, $\gamma$ and $\lambda$ are hyperparameters, $K$ is the number of leaves in $\tau_t$, and $\omega_{t,k}$ denotes the weight of its $k$-th leaf, $k=1,2,\dots,K$. By Equation~\eqref{eq:xgboost}, we have ${\theta}_i^{(t-1)}=\sum_{j=1}^{(t-1)}\tau_{j}(\x_i)$, $p_i^{(t-1)}=\sigma({\theta}_i^{(t-1)})$, and   
   \begin{equation}
	\label{eq:log_loss}
	l(y_i,\theta_i^{(t-1)})=-y_i\log(p_i^{(t-1)})-(1-y_i)\log(1-p_i^{(t-1)})=-y_i\theta_i^{(t-1)}+\log(1+e^{\theta_i^{(t-1)}}).
\end{equation}
Terms $g_{t,i}$ and $h_{t,i}$ in Equation \eqref{eq:xgboost_loss} are the first and second order partial derivatives of $l(y_i,\theta_i^{(t-1)})$ w.r.t. $\theta_i^{(t-1)}$, respectively. We have
\begin{equation}
	\label{eq:gi}
	g_{t,i}=\frac{\partial l}{\partial \theta_i^{(t-1)}}=\frac{1}{1+e^{-\theta_i^{(t-1)}}}-y_i=p^{(t-1)}_i-y_i,
\end{equation}
and 
\begin{equation}
	\label{eq:hi}
	h_{t,i}=\frac{\partial^2 l}{\partial^2 \theta_i^{(t-1)}}=\frac{e^{-\theta_i^{(t-1)}}}{(1+e^{-\theta_i^{(t-1)}})^2}=p^{(t-1)}_i(1-p^{(t-1)}_i).
\end{equation}
We define vectors $\g_t=(g_{t,1},\dots,g_{t,n})^{\prime}$ and $\h_t=(h_{t,1},\dots,h_{t,n})^{\prime}$, where notation $^\prime$ denotes vector transpose.

To construct tree $\tau_t$, XGBoost grows nodes by finding the best split to divide current training instances, until reaching the user-specified maximum depth. A candidate split $S$ is a combination of a focal feature and its associated threshold; current training instances are divided into two groups depending on whether their focal features are less than the threshold. To minimize the loss function~\eqref{eq:xgboost_loss}, the best split is the one achieving the maximum loss reduction among all candidate splits, where the loss reduction $\loss_{t,S}$ of a candidate split $S$ is given by:
\begin{align}
	\begin{aligned}
		\label{eq:loss_reduction}
		\loss_{t,S}&=\frac{1}{2}[\frac{(\g_t^{\prime}{I^L_{S}})^2}{\h_t^{\prime}{I^L_{S}}+\lambda}+\frac{(\g_t^{\prime}{I^R_{S}})^2}{\h_t^{\prime}{I^R_{S}}+\lambda}-\frac{(\g_t^{\prime}{I})^2}{\h_t^{\prime}{I}+\lambda}]-\gamma\\
		&=\frac{1}{2}[\frac{(G^L_{t,S})^2}{H^L_{t,S}+\lambda}+\frac{(G^R_{t,S})^2}{H^R_{t,S}+\lambda}-\frac{(G_t)^2}{H_t+\lambda}]-\gamma,
	\end{aligned}
\end{align}
In Equation~\eqref{eq:loss_reduction}, vector $I\in \{1,0\}^{n}$ denotes the set of current training instances; its $i$-th element $I_i=1$ means that instance $i$ is in the set and $I_i=0$ otherwise. Instances in $I$ are divided by $S$ into two subsets: $I^L_{S}$ and $I^R_{S}$, which respectively represent the set of current training instances with their focal features less than (moving to the left branch of the tree) and not less than (moving to the right branch of the tree)     
the threshold of $S$.
To simplify notation, we denote $G^L_{t,S}$ as the dot product of $\g_t$ and ${I^L_{S}}$, i.e., $G^L_{t,S}=\g_t^{\prime}{I^L_{S}}$. Similarly, $G^R_{t,S}=\g_t^{\prime}{I^R_{S}}$, $G_{t}=\g_t^{\prime}I$,
$H^L_{t,S}=\h_t^{\prime}{I^L_{S}}$, $H^R_{t,S}=\h_t^{\prime}{I^R_{S}}$, and $H_{t}=\h_t^{\prime}I$. Since $G_t$ and $H_t$ are fixed for all candidate splits, comparing loss reduction $\loss_{t,S}$ is equivalent to comparing $score_{t,S}$ defined below:
\begin{align}
	\begin{aligned}
		\label{eq:score_s}
		score_{t,S}&=\frac{(G^L_{t,S})^2}{H^L_{t,S}+\lambda}+\frac{(G^R_{t,S})^2}{H^R_{t,S}+\lambda}.
	\end{aligned}
\end{align} 
Thus, the best split is the one achieving the maximum $score_{t,S}$ among all candidate splits. Finally, the weight of tree $\tau_t$'s $k$-th leaf, $\omega_{t,k}$, is computed as:
\begin{equation}
	\label{eq:fitted_value}
	\omega_{t,k}=-\frac{\g_t^{\prime}I_t^k}{\h_t^{\prime}I_t^k+\lambda}=-\frac{G_t^k}{H_t^k+\lambda}.
\end{equation}
where vector $I_t^k\in\{1,0\}^{n}$ denotes the set of instances in the leaf, its $i$-th element $I_{i,t}^k=1$ means that instance $i$ is in the leaf and $I_{i,t}^k=0$ otherwise, $G^k_{t}=\g_t^{\prime}{I_t^k}$, and $H^k_{t}=\h_t^{\prime}{I_t^k}$. Interested readers please refer to \citep{chen2016xgboost} for the derivations of Equations \eqref{eq:loss_reduction} and \eqref{eq:fitted_value}.

\subsection{Paillier Additively Homomorphic Encryption}
\label{sec:background:paillier}

Proposed by \cite{paillier1999public}, Paillier additively homomorphic encryption, hereafter Paillier encryption, is an encryption scheme widely used by privacy-preserving machine learning methods \citep{cheng2021secureboost,liu2020federated}. In Paillier encryption, plaintexts refer to numerical and textual data that are not encrypted and ciphertexts refer to those encrypted. We illustrate Paillier encryption in the context of our study. To implement Paillier encryption, a pair of keys, namely public key and private key, are created by a trusted party (e.g., a government authority). Plaintexts can be encrypted with the public key but ciphertexts can only be decrypted with the private key. The trusted party only gives public key to the financial institution $F$ and the alternative data owner $A$. Before sending its alternative data to $F$, $A$ encrypts these data with public key. Since $F$ does not have private key, the encrypted data sent by $A$ cannot be decrypted by $F$, thereby ensuring no information leaked from $A$ to $F$. Similarly, $F$ encrypts its own data before sending them to $A$, which guarantees no information leaked from $F$ to $A$. 
Paillier encryption has several attractive properties. 

\textit{Non-Deterministic Encryption}: If the same plaintext is encrypted several times, its corresponding ciphertexts are different at different times. 

\noindent In addition, Paillier encryption allows certain algebraic operations to be conducted on ciphertexts without decrypting them, which are summarized in the following homomorphic properties \citep{paillier1999public}. Let $[[m]]$ denote the (encrypted) ciphertext of a plaintext $m$, e.g., plaintext $47$ and its ciphertext $[[47]]$.  

 \textit{Homomorphic Addition}: Given ciphertexts $[[m_1]]$ and $[[m_2]]$, we have 
 \begin{equation}
 	[[m_1+m_2]]=[[m_1]]\times[[m_2]]. \label{eq:pai_add}
 \end{equation}
\noindent According to this property, if a party receives ciphertexts $[[m_1]]$ and $[[m_2]]$, it can obtain the ciphertext of $m_1+m_2$, i.e., $[[m_1+m_2]]$, although it does not know $m_1$, $m_2$, nor $m_1+m_2$.

\textit{Homomorphic Subtraction}: Given ciphertexts $[[m_1]]$ and $[[m_2]]$, we have 
 \begin{equation}
 	[[m_1-m_2]]=[[m_1]]\times[[-m_2]]. \label{eq:pai_sub}
 \end{equation}

\textit{Homomorphic Scalar Multiplication}: Given plaintext $m_1$ and ciphertext $[[m_2]]$ , we have
\begin{equation}
[[m_1\times m_2]]=[[m_2]]^{m_1}.  \label{eq:pai_mul}
\end{equation}

\textit{Homomorphic Dot Product}: Given plaintext vector $\x$ and ciphertext vector $[[\z]]$, we can obtain the ciphertext of their dot product by
\begin{equation}
	[[\x^{\prime} \z]]=\prod_i [[z_{i}]]^{x_{i}}, \label{eq:pai_dot}
\end{equation}
 where $x_i$ and $[[z_i]]$ are elements of $\x$ and $[[\z]]$, respectively.

\section{Method}
\label{sec:method}

To solve the PCRPA problem, we propose a novel privacy-preserving credit risk prediction method, namely PrivacyCredit, which operates in the Paillier encryption environment. In this environment, a trusted authority $C$ generates a pair of public and private keys but only sends the public key to $F$ and $A$. As a result, both $F$ and $A$ can encrypt data with public key, but neither $F$ nor $A$ can decrypt encrypted data. Throughout this section, we denote encrypted data using notation $[[.]]$. PrivacyCredit consists of two building blocks, each of which addresses one methodological challenge of the PCRPA problem. We describe these two building blocks in Sections \ref{sec:method:training} and \ref{sec:method:inference}, respectively.

\subsection{Privacy-preserving Learning of Credit Risk Prediction Model}
Given training data that consist of traditional data ($X^F$, $\bd y$) at $F$ and alternative data $X^A$ at $A$, we propose how to learn the credit risk prediction model $\M$ from the training data while satisfying the three constraints of the PCRPA problem. As discussed in Section \ref{sec:background:xgboost}, $\M$ consists of a series of regression trees. Accordingly, we show how to learn a regression tree in a privacy-preserving manner in Section \ref{sec:method:training:regression tree}. Once a regression tree is learned, parameters $\g_t$ and $\h_t$ need to be updated for the construction of the next tree. Thus, we propose how to update these parameters in a privacy-preserving manner in Section \ref{sec:method:training:update}.

\label{sec:method:training}
\subsubsection{Privacy-preserving Learning of Regression Tree.}
\label{sec:method:training:regression tree}

Since the learning process is the same for all regression trees, we drop tree number subscript $t$ in this subsection for  simplicity of notation.
Like regular XGBoost \citep{chen2016xgboost}, PrivacyCredit first specifies all candidate splits before model training. Recall that a candidate split is a combination of a feature and its associated threshold. In particular, $F$ defines a set of candidate splits, denoted by $\s^F$; each split is defined on a traditional feature in $X^F$. 
Similarly, $A$ specifies a set of candidate splits, denoted by $\s^A$, each defined on an alternative feature in $X^A$.
We denote all candidate splits $\s=\s^F \cup \s^A$. For each $S\in\s$, we define a vector $I^<_{S}\in\{1,0\}^{n}$, where the $i$-th entry equals 1 if the focal feature value of training instance $i$ is less than the threshold associated with $S$, and 0 otherwise.

\begin{example}
Consider a data set of six training instances, whose annual incomes are $\$100K$, $\$200K$, $\$300K$, $\$150K$, $\$180K$, and $\$205K$, respectively. Given a candidate split $S$ = (feature = Annual Income, threshold = $\$200K$), we have $I^<_{S}=(1,0,0,1,1,0)^\prime$, because annual incomes of instances 1, 4, and 5 are less than $\$200K$.
\end{example}

Let vector $I\in\{1,0\}^{n}$ denote the set of current training instances.\footnote{Set $I$ contains all instances in the training data at the root node, but contains a subset of these instances at an intermediate node.} 
By applying a split $S$ on $I$, current instances are divided into two subsets: those with focal feature values less than the threshold of $S$, denoted by vector $I^L_{S}$, and those with focal feature values not less than the threshold of $S$, represented by vector $I^R_{S}$. 

\begin{example}
Continue with Example 1. Consider $I=(0,0,1,1,1,0)^\prime$; that is, current training instances consist of instances 3, 4, and 5.  Applying split $S$ = (feature = Annual Income, threshold = $\$200K$) to $I$,  we have $I^L_{S}=(0,0,0,1,1,0)^\prime$ and $I^R_{S}=(0,0,1,0,0,0)^\prime$, as instances 4 and 5 have annual incomes less than the threshold and instance 3 has annual income greater than it. 
\end{example}

Vector $I^L_{S}$ can be computed by 
\begin{equation}
\label{eq:I_relation}
I^L_{S} = I \wedge I^<_{S} ,
\end{equation}
where $\wedge$ is the element-wise logical conjunction operator. Equation~\eqref{eq:I_relation} can be easily verified using $I^L_{S}, I,$ and $I^<_{S}$ in the previous examples.

Having introduced $\s$ and $I^<_{S}$, we next describe how PrivacyCredit constructs a regression tree in a privacy-preserving manner. Specifically, it grows a node of the tree by finding the best split that achieves the maximum score defined by Equation~\eqref{eq:score_s}. According to the equation, the score of a split $S$ depends on  $\g=(g_1,g_2,\dots,g_n)^\prime$, $\h=(h_1,h_2,\dots,h_n)^\prime$, $I^L_{S}$, and $I^R_{S}$. To build a tree in a privacy-preserving manner, we need to protect the following sources of privacy leakage associated with $\g$, $\h$, $I^L_{S}$, and $I^R_{S}$:

(1) User $v_i$'s credit status $y_i$ can be inferred from $g_i$ and $h_i$, $i=1,2,\dots,n$. For example, according to Equation~\eqref{eq:gi}, $g_i=p_i-y_i$; thus, one can infer $y_i=1$ if $g_i<0$ and $y_i=0$ otherwise.

(2) If a user belongs to $I^L_{S}$, one can infer that the focal feature value of that user is less than the threshold of $S$. For example, given $S$ = (feature = Annual Income, threshold = $\$200K$), a user in $I^L_{S}$ indicates that the user's annual income is less than $\$200K$. Similarly, if a user belongs to $I^R_{S}$, one can infer that the focal feature value of the user is not less than the threshold of $S$.

To protect the first source of privacy leakage and prevent the leakage of credit status information from $F$ to $A$, $\g$ and $\h$ must be stored at $F$ and should not be transferred to $A$ without encryption. To address the second source of privacy leakage, $I^L_{S}$ and  $I^R_{S}$ must be encrypted before transferring from one party to the other. By Equation~\eqref{eq:I_relation}, $I^L_{S}$ is computed from $I^<_{S}$ and $I$. Similar to $I^L_{S}$, if a user belongs to $I^<_{S}$, one can infer that the focal feature value of the user is less than the threshold of $S$. Thus, $I^<_{S}$ should be encrypted before transferring from one party to the other. Moreover, current level training instances $I$ are either instances in $I^L_{S}$ or instances in $I^R_{S}$, which result from the split of the instances at one level higher using $S$. Therefore, $I$ should also be encrypted. 

We first show how $F$ obtains $[[I^<_{S}]]$, $\forall S \in \s$, by invoking Algorithm \ref{alg:compare_share}, where $X^A$ and $X^F$ denote data held by $A$ and $F$, respectively. As illustrated, $A$ derives and encrypts $I^<_{S}$ for each $S\in\s^A$ by comparing the focal feature value of each instance in $X^A$ and the threshold of $S$.  Since the credit risk prediction model is learned and stored at $F$, $A$ sends $[[I^<_{S}]]$ to $F$, $\forall S\in \s^A$. Finally, $F$ derives and encrypts $I^<_{S}$ for each $S\in\s^F$ by comparing the focal feature value of each instance in $X^F$ and the threshold of $S$. This way, $F$ obtains $[[I^<_{S}]], \forall S \in \s$, without violating the privacy-preserving constraint. 

\OneAndAHalfSpacedXI 
\begin{algorithm}
	\caption{$F$ Obtaining $[[I^<_{S}]]$}
	\label{alg:compare_share}
	\textbf{Input}:\quad $A$: $X^A$, $\s^A$; \quad $F$: $X^F$, $\s^F$;\\
	\textbf{Output}:\quad$F$: $[[I^<_{S}]]$, $\forall S\in\s$.
	\begin{algorithmic}[1]
		\STATE{ $A$ derives and encrypts $I^<_{S}$, for each $S\in\s^A$.}
		\STATE{ $A$ sends $[[I^<_{S}]]$ to $F$, $\forall S\in \s^A$.}
		\STATE{$F$ derives and encrypts $I^<_{S}$, for each $S\in\s^F$.}
	\end{algorithmic}
\end{algorithm} 
\DoubleSpacedXI

Next, we show how to compute encrypted $[[I^L_{S}]]$ and $[[I^R_{S}]]$, given $S$ and encrypted $[[I]]$ and $[[I^<_{S}]]$. If we can replace the element-wise logical conjunction operator in Equation~\eqref{eq:I_relation} with an element-wise addition operator, we can employ the homomorphic addition property (i.e., Equation \eqref{eq:pai_add}) to derive $[[I^L_{S}]]$. Let $I^L_{i,S}$, $I_i$, and $I^<_{i,S}$ denote an element of $I^L_S$, $I$, and $I^<_S$, respectively. We have $I^L_{i,S} = I_i \wedge I^<_{i,S} = 1$ if and only if $I_i + I^<_{i,S} = 2$ (i.e., both $I_i$ and $I^<_{i,S}$ being 1) and   $I^L_{i,S} = I_i \wedge I^<_{i,S} = 0$ if and only if $I_i + I^<_{i,S} < 2$ (i.e., at least one of $I_i$ and $I^<_{i,S}$ not being 1). Therefore, we can derive an element $[[I^L_{i,S}]]$ of $[[I^L_S]]$ by computing $[[I_i + I^<_{i,S}]]$ with the homomorphic addition property:
\begin{equation*}
[[I_i + I^<_{i,S}]] = [[I_i]] \times[[I^<_{i,S}]],
\end{equation*}
and then comparing $[[I_i + I^<_{i,S}]]$ and $[[2]]$ using the encrypted data comparison scheme \citep{veugen2011comparing}.\footnote{ The encrypted data comparison scheme takes two ciphertexts as input and outputs comparison results in the forms of $[[1]]$ or $[[0]]$.} An element $[[I^R_{i,S}]]$ of $[[I^R_S]]$ can be derived with the homomorphic subtraction property (i.e., Equation \eqref{eq:pai_sub}):  
\begin{equation*}
[[I^R_{i,S}]] = [[I_i - I^L_{i,S}]] = [[I_i]] \times[[-I^L_{i,S}]].
\end{equation*}

With $[[I^L_{S}]]$ and $[[I^R_{S}]]$ derived as well as $\g$ and $\h$ on hand, $F$ needs to compute encrypted $[[score_S]]$ according to Equation \eqref{eq:score_s}.
Specifically, encrypted term $[[G^L_S]]$ of the equation can be calculated as $[[G^L_S]]=[[\g^{\prime}I^L_{S}]]$, which can be further computed using the homomorphic dot product property (i.e., Equation \eqref{eq:pai_dot}). Using the same property, $F$ can compute encrypted terms $[[G^R_S]]=[[\g^{\prime}I^R_{S}]]$, $[[H^L_S]]=[[\h^{\prime}I^L_{S}]]$, and $[[H^R_S]]=[[\h^{\prime}I^R_{S}]]$. The difficulty in computing $[[score_S]]$ with Equation \eqref{eq:score_s} is that division over encrypted terms is not defined in the Paillier encryption scheme \citep{paillier1999public}. We propose the following procedure and Lemma \ref{lma:score_s} to tackle this difficulty.
Specifically, $F$ generates four random numbers $\alpha_1$, $\alpha_2$, $\beta_1$, $\beta_2$, and sends $[[N_1]]=[[G^L_S+\alpha_1]]$, $[[N_2]]=[[G^R_S+\alpha_2]]$, $[[M_1]]=[[\beta_1(H^L_S+\lambda)]]$, and $[[M_2]]=[[\beta_2(H^R_S+\lambda)]]$ to the trusted authority $C$.\footnote{By the homomorphic addition property (i.e., Equation \eqref{eq:pai_add}), $[[N_1]]=[[G^L_S+\alpha_1]]=[[G^L_S]]\times[[\alpha_1]]$. Using the same property, we can compute $[[N_2]]$. $[[M_1]]$ and $[[M_2]]$ can be computed using the homomorphic addition property (i.e., Equation \eqref{eq:pai_add}) and the homomorphic scalar multiplication property (i.e., Equation \eqref{eq:pai_mul}).} Since $C$ has private key, $C$ can decrypt them and obtain plaintexts $N_1$, $N_2$, $M_1$, and $M_2$. 
It is noted that $C$ cannot infer $G^L_S$ from $N_1$, $G^R_S$ from $N_2$, $H^L_S$ from $M_1$, and $H^R_S$ from $M_2$ because $C$ does not know random numbers $\alpha_1$, $\alpha_2$, $\beta_1$, $\beta_2$.\footnote{ The secrecy of the element-wise random mask is guaranteed based on the inability to definitively solve $n$ equations with more than $n$ unknowns \citep{du2004privacy,vaidya2002privacy}.} Next, $C$ computes and sends encrypted 
$[[\frac{(N_1)^2}{M_1}]]$,  $[[\frac{(N_2)^2}{M_2}]]$,  $[[\frac{N_1}{M_1}]]$, $[[\frac{N_2}{M_2}]]$, $[[\frac{1}{M_1}]]$, and $[[\frac{1}{M_2}]]$ to $F$, and   
$F$ calculates $[[score_S]]$ by Lemma~\ref{lma:score_s}.
\begin{lemma}
	\label{lma:score_s}
	The encrypted score of a candidate split $S$, $[[score_S]]$, can be calculated as:
	\begin{align*}
		[[score_S]] = &\quad [[\frac{(N_1)^2}{M_1}]]^{\beta_1}\times [[-\frac{N_1}{M_1}]]^{2\alpha_1\beta_1}\times[[\frac{1}{M_1}]]^{\alpha_1^2\beta_1}\\
		&\times[[\frac{(N_2)^2}{M_2}]]^{\beta_2}\times[[-\frac{N_2}{M_2}]]^{2\alpha_2\beta_2}\times[[\frac{1}{M_2}]]^{ \alpha_2^2\beta_2}.
	\end{align*}
	\textit{Proof.} See Appendix~\ref{apd:lma}.
\end{lemma}

Using the procedure described above, $F$ computes the encrypted score for each candidate split. It then finds the best split $S^*$ with the maximum score as well as its corresponding $[[I^L_{S^*}]]$ and $[[I^R_{S^*}]]$, by applying the \textit{argmax} over encrypted data scheme proposed by \citet{bost2015machine} to all encrypted scores.\footnote{ \textit{Argmax} over encrypted data scheme takes encrypted numbers as input and outputs the position of the largest number in plaintext. } 
As a result, only $F$ knows the best split $S^*$ and the model-confidential constraint is maintained. Algorithm~\ref{alg:split_finding} summarizes the privacy-preserving finding of the best split. 
 
\OneAndAHalfSpacedXI
\begin{algorithm}
	\caption{Privacy-Preserving Finding of Best Split}
	\label{alg:split_finding}
	\textbf{Input}:\quad $F$: $\lambda$, $\g$, $\h$, $[[I]]$, $[[I^<_{S}]]$ $\forall S\in\s$\\
	\textbf{Output}:\quad$F$: $S^*$, $[[I^L_{S^*}]]$, $[[I^R_{S^*}]]$
	\begin{algorithmic}[1]
		\FOR{each $S\in \s$} 
		\STATE{$F$ derives $[[I^L_{S}]]$ and $[[I^R_S]]$ from $[[I]]$ and $[[I_S^<]]$.}
		\STATE{$F$ computes $[[G^L_S]]$, $[[G^R_S]]$, $[[H^L_S]]$, and $[[H^R_S]]$ by applying Equation \eqref{eq:pai_dot}.}
		\STATE{$F$ generates random numbers $\alpha_1$, $\alpha_2$, $\beta_1$, $\beta_2$, and sends $[[N_1]]=[[G^L_S+\alpha_1]]$, $[[N_2]]=[[G^R_S+\alpha_2]]$, $[[M_1]]=[[\beta_1(H^L_S+\lambda)]]$, and $[[M_2]]=[[\beta_2(H^R_S+\lambda)]]$ to $C$.}
		\STATE{$C$ decrypts $[[N_1]], [[N_2]],[[M_1]],[[M_2]]$ and sends encrypted 
			$[[\frac{(N_1)^2}{M_1}]]$,  $[[\frac{(N_2)^2}{M_2}]]$,  $[[\frac{N_1}{M_1}]]$, $[[\frac{N_2}{M_2}]]$, $[[\frac{1}{M_1}]]$, and $[[\frac{1}{M_2}]]$ to $F$.} 
		\STATE{$F$ calculates $[[score_S]]$ by Lemma~\ref{lma:score_s}.}
		\ENDFOR
		\STATE{$F$ obtains the best split $S^*$ as well as its corresponding $[[I^L_{S^*}]]$ and $[[I^R_{S^*}]]$ by applying the \textit{argmax} over encrypted data scheme to all encrypted scores.}
	\end{algorithmic}
\end{algorithm}
\DoubleSpacedXI

To construct a regression tree, Algorithm~\ref{alg:split_finding} is iteratively invoked to find the best split that divides current training instances, until reaching the user-specified maximum depth of the tree. Finally, $F$ computes the encrypted weight for each leaf of the tree according to Equation~\eqref{eq:fitted_value}. Again, the challenge in computing Equation \eqref{eq:fitted_value} is that division over encrypted terms is not defined in the Paillier encryption scheme. 
To address this challenge and derive the encrypted weight of the $k$-th leaf, $[[\omega_k]]$, $F$ generates random numbers $\alpha_k$ and $\beta_k$ and sends $[[N_k]]=[[G^k+\alpha_k]]$ and $[[M_k]]=[[\beta_k(H^k+\lambda)]]$ to $C$, where $[[G^k]]=[[\g^\prime I^k]]$, $[[H^k]]=[[\h^\prime I^k]]$, and $I^k\in\{1,0\}^{n}$ denotes the set of instances in the leaf. $C$ decrypts them and sends encrypted $[[-\frac{N_k}{M_k}]]$ and $[[-\frac{1}{M_k}]]$ to $F$, which in turn calculates $[[\omega_k]]$ by Lemma~\ref{lma:weight}.
\begin{lemma}
	\label{lma:weight}
	The encrypted weight of the $k$-th leaf, $[[\omega_k]]$ can be calculated as:
	\begin{equation*}
	[[\omega_k]] = [[-\frac{N_k}{M_k}]]^{\beta_k}\times[[\frac{1}{M_k}]]^{\alpha_k\beta_k}.
	\end{equation*}
    \textit{Proof.} See Appendix~\ref{apd:lma}.
\end{lemma}

\subsubsection{Privacy-preserving Parameter Updating.}
\label{sec:method:training:update}

After building the $t$-th tree $\tau_t$, PrivacyCredit updates $\g_t$ to $\g_{t+1}$ and  $\h_t$ to $\h_{t+1}$ for the construction of the $(t+1)$-th tree $\tau_{t+1}$. According to Equations~\eqref{eq:gi} and \eqref{eq:hi}, both $g_{t+1,i}$ and $h_{t+1,i}$ can be derived from $p_i^{(t)}$, $i=1,2,\cdots,n$. It is thus imperative to compute $p_i^{(t)}$. 
By Equation~\eqref{eq:xgboost}, we have  
\begin{align}
	\begin{aligned}
		\label{eq:xgboost_rec}
		p_i^{(t)}&=\sigma({\theta}_i^{(t)}), \quad \text{and}\\
		\theta_i^{(t)} &= \theta_i^{(t-1)} + \tau_{t}(\x_i)\\
		&= \theta_i^{(t-1)} + \omega_{t,q_t(\x_i)}, \quad \text{by Equation \eqref{eq:tau_t}} 
	\end{aligned}
\end{align}
where $\x_i$ denotes user $v_i$'s traditional and alternative feature values. Recall that function $q_t$ takes $\x_i$ as input and classifies $v_i$ into one of the $K$ leaves of tree $\tau_t$. Therefore, we have 
\begin{equation}
	\label{eq:xgboost_sum}
	\omega_{t,q_t(\x_i)} = \sum_{k=1}^{K} \omega_{t,k} I_{i,t}^k,
\end{equation}
where $I_{i,t}^k$ is the $i$-th element of $I_t^k$, $I_{i,t}^k=1$ if $v_i$ belongs to the $k$-th leaf of $\tau_t$, and $I_{i,t}^k=0$ otherwise. Note that $F$ only has encrypted $[[\omega_{t,k}]]$ and $[[I_{t}^k]]$. To calculate $\omega_{t,q_t(\x_i)}$ from encrypted $[[\omega_{t,k}]]$ and $[[I_{t}^k]]$, we propose the following procedure. First, $F$ generates random numbers $\alpha_k$ and $\beta_k$, and sends $[[W_k]]=[[\omega_{t,k}+\alpha_k]]$ and $[[O_k]]=[[I_{i,t}^k+\beta_k]]$ to $C$, $k= 1,2,\dots,K$. Next, $C$ deciphers them and sends encrypted $[[W_k\times O_k]]$ to $F$, $k= 1,2,\dots,K$. By Lemma~\ref{lma:update}, $F$  calculates $[[\omega_{t,q_t(\x_i)}]]$ from $[[W_k\times O_k]]$. To get plaintext $\omega_{t,q_t(\x_i)}$, $F$ generates another random number $\xi$ and sends $[[\Upsilon]]=[[\omega_{t,q_t(\x_i)}+\xi]]$ to $C$. Finally, $C$ decrypts $[[\Upsilon]]$ and sends $\Upsilon$ to $F$, and $F$ obtains $\omega_{t,q_t(\x_i)}$ as $\Upsilon-\xi$. Algorithm~\ref{alg:update} summarizes the procedure of computing $p_i^{(t)}$. Once $p_i^{(t)}$ is derived, $F$ can compute $g_{t+1,i}$ and $h_{t+1,i}$ according to Equations~\eqref{eq:gi} and \eqref{eq:hi} respectively, $i=1,2,\cdots,n$.

\OneAndAHalfSpacedXI
\begin{algorithm}
	\caption{Privacy-Preserving Computation of $p_i^{(t)}$}
	\label{alg:update}
	\textbf{Input}:\quad $F$: $[[\omega_{t,k}]]$, $[[I_{i,t}^k]]$, $k=1,2,\dots,K$\\
	\textbf{Output}:\quad$F$: $p_i^{(t)}$
		\begin{algorithmic}[1]
		\STATE{$F$ generates random numbers $\alpha_k$ and $\beta_k$, and sends $[[W_k]]=[[\omega_{t,k}+\alpha_k]]$ and $[[O_k]]=[[I_{i,t}^k+\beta_k]]$ to $C$, $k=1,2,\dots,K$.}
		\STATE{$C$ decrypts $[[W_k]]$, $[[O_k]]$ and sends encrypted $[[W_k\times O_k]]$ to $F$,  $k=1,2,\dots,K$.}
		\STATE{$F$ calculates $[[\omega_{t,q_t(\x_i)}]]$ by Lemma~\ref{lma:update}}.
		\STATE{$F$ generates a random number  $\xi$ and sends $[[\Upsilon]]=[[\omega_{t,q_t(\x_i)}+\xi]]$ to $C$.}
		\STATE{$C$  decrypts $[[\Upsilon]]$ and sends $\Upsilon$ to $F$.}
		\STATE{$F$  calculates $\omega_{t,q_t(\x_i)}=\Upsilon-\xi$ and computes $p_i^{(t)}$ by Equation~\eqref{eq:xgboost_rec}.}
	\end{algorithmic}
\end{algorithm} 
\DoubleSpacedXI

\begin{lemma}
	\label{lma:update}
	Encrypted $[[\omega_{t,q_t(\x_i)}]]$ can be calculated as:
	\begin{equation*}
		[[\omega_{t,q_t(\x_i)}]]=\prod_k([[W_k\times O_k]]\times[[-I_{i,t}^k]]^{\alpha_k}\times[[-\omega_{t,k}]]^{\beta_k}\times[[-\alpha_k\beta_k]]).
	\end{equation*}
	\textit{Proof.} See Appendix~\ref{apd:lma}.
\end{lemma}

\subsection{Privacy-preserving Credit Risk Prediction}
\label{sec:method:inference}

The learned credit risk prediction model can be employed to predict the credit risk of a new user. Let vector $\x_{new}^F\in\real^{d^F}$ denote the new user's traditional feature values stored at $F$ and vector $\x_{new}^A\in\real^{d^A}$ be the user's alternative feature values held by $A$. Since the model is stored at $F$, $A$ sends encrypted $[[\x_{new}^A]]$ to $F$. We denote the new user's data as vector $\x_{new}$, which is the concatenation of $\x_{new}^F$ and $[[\x_{new}^A]]$. Each tree $\tau_t$ of the model takes $\x_{new}$ as the input and returns encrypted $[[\omega_{t,q_t(\x_{new})}]]$, where function $q_t$ classifies the new user into one of the $K$ leaves of the tree and $t=1,2,\dots,T$. Similar to Equation \eqref{eq:xgboost_sum}, we have $[[\omega_{t,q_t(\x_{new})}]]=[[\sum_{k=1}^{K} \omega_{t,k} \delta^k_t]]$, where $\omega_{t,k}$ is the weight of the $k$-th leaf of $\tau_t$, $\delta^k_t\in\{1,0\}$ denotes whether the new user is classified into the $k$-th leaf of $\tau_t$, and $k=1,2,\dots,K$.

For each branch $b$ of $\tau_t$, let $\delta_{b}=1$ denote the new user being classified along the branch and $\delta_{b}=0$ otherwise. The value of $\delta_{b}$ depends on the split attribute (e.g., number of late payments of phone bills), threshold (e.g., 5), and comparison operator (e.g., greater than) associated with branch $b$ as well as the new user's corresponding attribute value. The value of $\delta_{b}$ should be encrypted. For example, if $\delta_{b}=1$ is disclosed, one can infer that the new user's number of late payments of phone bills is greater than 5. To compute $[[\delta_{b}]]$ for each branch $b$ of $\tau_t$, $F$ compares the threshold associated with the branch and the new user's corresponding attribute value using the encrypted data comparison scheme \citep{veugen2011comparing}.

Let $Path_{t,k}$ denote the path from the root node to the $k$-th leaf of $\tau_t$, $k=1,2,\dots,K$, which consists of a sequence of branches in the path. In plaintext, if the summation of $\delta_{b}$s of these branches equals the depth of $\tau_t$, $\delta_{b}$ of each branch in the path must be $1$; thus, the new user is classified by following each branch in the path and we have $\delta^k_t=1$. On the other hand, if the summation of $\delta_{b}$ of these branches is less than the depth of $\tau_t$, $\delta_{b}$ of some (or all) branches in the path must be $0$ and we have $\delta^k_t=0$. However, $F$ has to work with encrypted $[[\delta_{b}]]$ and $[[\delta^k_t]]$ to satisfy the privacy-preserving constraint. To accomplish this, by the homomorphic addition property (i.e., Equation \eqref{eq:pai_add}), $F$ computes
\begin{equation}
\label{eq:path_sum_enc}
[[\sum_{b\in Path_{t,k}}\delta_{b}]]=\prod_{b\in Path_{t,k}}[[\delta_{b}]].
\end{equation}
Next, $F$ derives $[[\delta^k_t]]$ by comparing $[[\sum_{b\in Path_{t,k}}\delta_{b}]]$ and the depth of $\tau_t$ using the encrypted data comparison scheme \citep{veugen2011comparing}.

With $[[\delta^k_t]]$ derived for the new user and $[[\omega_{t,k}]]$ given by the model, $F$ generates random numbers $\alpha_k$ and $\beta_k$, and sends $[[W_k]]=[[\omega_{t,k}+\alpha_k]]$ and $[[O_k]]=[[\delta^k_t+\beta_k]]$ to $C$, $k= 1,2,\dots,K$. Party $C$ then deciphers them and sends encrypted $[[W_k\times O_k]]$ to $F$. By Lemma~\ref{lma:update}, $F$ calculates $[[\omega_{t,q_t(\x_{new})}]]$ as:
\begin{equation}
\label{eq:w_t_q}
[[\omega_{t,q_t(\x_{new})}]]=\prod_k([[W_k\times O_k]]\times[[-\delta^k_t]]^{\alpha_k}\times[[-\omega_{t,k}]]^{\beta_k}\times[[-\alpha_k\beta_k]]).
\end{equation}
After deriving  $[[\omega_{t,q_t(\x_{new})}]]$ for $t=1,2,\dots, T$, $F$ calculates $[[\theta_{new}]]=[[\sum_{t=1}^T\omega_{t,q_t(\x_{new})}]]=\prod_{t=1}^T[[\omega_{t,q_t(\x_{new})}]]$ by Equation \eqref{eq:xgboost} and the homomorphic addition property. To get plaintext $\theta_{new}$, $F$ sends $[[\Theta]]=[[\theta_{new}+\rho]]$ to $C$, where $\rho$ is a random number generated by $F$. Party $C$ deciphers $[[\Theta]]$ and sends $\Theta$ to $F$. Finally, $F$ obtains $\theta_{new}=\Theta-\rho$ and calculates the delinquent probability $p_{new}$ of the new user by Equation \eqref{eq:xgboost}, i.e., $p_{new}=\sigma(\theta_{new})$. We illustrate the procedure of privacy-preserving credit risk prediction in Algorithm~\ref{alg:Inf}.
\OneAndAHalfSpacedXI
\begin{algorithm}[H]
	\caption{Privacy-Preserving Credit Risk Prediction}
	\label{alg:Inf}
	\textbf{Input}: $F$: $\x_{new}^F$, $\tau_t$, $t=1,2,\dots,T$;\quad $A$: $\x_{new}^A$\\
	\textbf{Output}: $F$: $p_{new}$
	\begin{algorithmic}[1]
		\FOR{$t=1,2,\dots,T$}
		\STATE{$F$ computes $[[\delta_{b}]]$ for each branch $b$ in $\tau_t$.}
		\STATE{$F$ calculates $[[\sum_{b\in Path_{t,k}}\delta_{b}]]$ by Equation~\eqref{eq:path_sum_enc}, $k=1,2,\dots,K$. }
		\STATE{$F$ derives $[[\delta^k_t]]$ by comparing $[[\sum_{b\in Path_{t,k}}\delta_{b}]]$ and the depth of the $k$-th leaf with the encrypted data comparison scheme, $k=1,2,\dots,K$.}
		\STATE{$F$ generates random numbers $\alpha_k$ and $\beta_k$, and sends $[[W_k]]=[[\omega_{t,k}+\alpha_k]]$ and $[[O_k]]=[[\delta^k_t+\beta_k]]$ to $C$, $k=1,2,\dots,K$.}
		\STATE{$C$ decrypts $[[W_k]]$, $[[O_k]]$ and sends encrypted $[[W_k\times O_k]]$ to $F$,  $k=1,2,\dots,K$.}
		\STATE{$F$ calculates $[[\omega_{t,q_t(\x_{new})}]]$ by Equation \eqref{eq:w_t_q}}.
		
		\ENDFOR
		\STATE{$F$ computes $[[\theta_{new}]]=[[\sum_{t=1}^T\omega_{t,q_t(\x_{new})}]]=\prod_{t=1}^T[[\omega_{t,q_t(\x_{new})}]]$.}
		\STATE{ $F$ generates a random number $\rho$ and sends $[[\Theta]]=[[\theta_{new}+\rho]]$ to $C$.}
		\STATE{$C$ deciphers $[[\Theta]]$ and sends $\Theta$ to $F$.}
		\STATE{$F$ obtains $\theta_{new}=\Theta-\rho$ and computes $p_{new}=\sigma(\theta_{new})$.}
	\end{algorithmic}
\end{algorithm}
\DoubleSpacedXI

The following proposition shows that the proposed PrivacyCredit method satisfies all three constraints of the PCRPA problem.

\begin{proposition}[Properties of PrivacyCredit]
\label{thm:pcrpa_constraints}\rm
PrivacyCredit satisfies the privacy-preserving, model-confidential, and lossless constraints of the PCRPA problem.
\begin{enumerate}
    \item[(i)] \textbf{Privacy preservation:} During the model learning and inference procedures of PrivacyCredit, no traditional data $(X^F,\bd y)$ held by $F$ are disclosed to $A$, and no alternative data $X^A$ held by $A$ are disclosed to $F$.

    \item[(ii)] \textbf{Model confidentiality:} PrivacyCredit learns and stores the credit risk prediction model $\M$ centrally at $F$, ensuring that no other party can access any information about the model.

    \item[(iii)] \textbf{Losslessness:} The model $\M$ learned by PrivacyCredit achieves the same predictive performance as the credit risk prediction model $\M^{\circ}$ trained on the insecure plaintext combination of $(X^F,\bd y)$ and $X^A$.
\end{enumerate}

\proof{$\rm Proof.$}
\rm See Appendix~\ref{apd:thm}. \endproof
\end{proposition}

\subsection{Computational Complexity Analysis}
\label{subsec:complexity-analysis}

We then analyze the computational complexity of PrivacyCredit. Let $n$ denote the number of common users in the training set, $m=|\mathcal{S}_F\cup\mathcal{S}_A|$ be the total number of candidate splits, $T$ represent the number of boosting trees, and $D$ denote the maximum depth of these trees. If $q$ candidate thresholds are generated for each feature, then $m=O(q(d_F+d_A))$. We treat the bit lengths used in encrypted comparisons as fixed unless otherwise stated; the detailed primitive-level analysis is provided in Appendix~\ref{app:complexity}.

Algorithm~\ref{alg:compare_share} precomputes encrypted split indicators for all candidate splits. For each split, it constructs and encrypts an indicator vector of length $n$. Thus, Algorithm~\ref{alg:compare_share} requires $O(nm)$ encryptions. This step is performed once before boosting begins. Next, the leading-order term in the training complexity comes from Algorithm~\ref{alg:split_finding}, which is invoked at every internal node of every tree. At one internal node, PrivacyCredit evaluates all $m$ candidate splits. For each split, it constructs encrypted left/right child indicators and computes four encrypted sufficient statistics, corresponding to the gradient and Hessian sums on the two child nodes. These operations require scanning vectors of length $n$, so the per-split cost is linear in $n$. In addition, selecting the best split requires an encrypted \textit{argmax} over $m$ encrypted scores. Hence, the cost of one internal node is $ O(mn + m\ell_s)$, where $\ell_s$ is the bit length of the fixed-point encoded split scores used in the encrypted \textit{argmax} by \citet{bost2015machine}. Moreover, a full binary tree of depth $D$ contains $O(2^D)$ internal and leaf nodes. As a result, constructing one tree costs $O\left(2^D[m(n+\ell_s)+n]\right)$, where the final $n$ term captures encrypted leaf-weight computation and prediction updates over the training users. Across $T$ boosting rounds, the total training complexity is $O(nm) + O\left(T2^D[m(n+\ell_s)+n]\right)$. When $\ell_s$ is treated as a fixed precision parameter, we have $m(n+\ell_s)+n = mn + O(m) + n$. Since $m\geq 1$ and $n\geq 1$, the lower-order terms $O(m)$ and $n$ are dominated by $mn$. Moreover, the one-time precomputation term $O(nm)$ is dominated by the boosting-stage term when $T2^D\geq 1$. Therefore, the total training complexity simplifies to $O(T2^Dmn)$.
For inference, predicting a new borrower does not require scanning the training users or evaluating all candidate splits. Instead, PrivacyCredit computes encrypted branch and leaf indicators for each learned tree and combines encrypted leaf weights. Thus, the inference complexity for one borrower is $O(T2^D)$ for fixed comparison bit lengths.

The resulting complexity expression has several implications. First, the training cost scales linearly with the number of users $n$, since each candidate split requires scanning encrypted indicator vectors of length $n$. Second, the training cost scales linearly with the number of candidate splits $m$, implying a direct tradeoff between split granularity and computational efficiency. Third, the cost scales linearly with the number of boosting trees $T$. Finally, both training and inference scale exponentially with the maximum tree depth $D$, because the number of tree nodes grows as $2^D$.

\section{Empirical Evaluation}
\label{sec:eval}

\providecommand{\tbd}{\textcolor{red}{TBD}}

\subsection{Data and Evaluation Procedure}
\label{sec:eval:data_eval}

We collected traditional credit data $(X^F,\bd{y})$ from a financial institution in an Asian country, which offers a digital-payment credit product through an online platform. The matrix $X^F$ contains credit-related data for $11,992$ users of this product; each row of $X^F$ describes a user with seven features, such as the average number of transactions per month, average statement balance, and number of historical delinquencies. A user's credit status $y_i\in \bd y$ is set to $1$ (i.e., delinquency) if the user is more than 90 days past due on the minimum payment, and to $0$ (i.e., no delinquency) otherwise. Among the $11,992$ users, the delinquency rate is $8.59\%$. 
We also obtained alternative data $X^A$ for these $11,992$ users; these data were originally provided by a mobile service provider and made available to us through the financial institution.\footnote{For users who successfully obtain the digital-payment credit product, the product agreement authorizes the financial institution to collect and use relevant alternative data for  credit risk monitoring. Such alternative data include mobile usage data provided by a mobile service provider.} $X^A$ includes five characteristics of mobile usage, such as the number of late phone bill payments, number of contacts, and average daily call duration.\footnote{Both the credit data and alternative data were de-identified and used solely for research purposes.}

We next detail the evaluation procedure. Our method was implemented in a privacy-preserving environment.  Specifically, we used three separate desktops to simulate the three participating parties: the financial institution $F$, which stored the credit data $(X^F,\bd{y})$; the alternative data provider $A$, which stored the alternative mobile data $X^A$; and the trusted authority $C$, which held the secret decryption key. All intermediate results exchanged among the three desktops were encrypted or randomized to satisfy the privacy-preserving constraint, and all information about the learned credit risk prediction model was retained only on the desktop representing the financial institution $F$. As a result, PrivacyCredit learned a credit risk prediction model from both $(X^F,\bd{y})$ and $X^A$ while preserving the privacy of the alternative data and the confidentiality of the learned model.

We employed 70\% of users for training and the remaining 30\% for testing. Using the training data, we trained PrivacyCredit and each benchmark method to predict the delinquency probability of each user in the test data.
The predictive performance is evaluated using four commonly used metrics for credit risk prediction: precision, recall, F1 score, and AUC \citep{lu2023profit,zhang2025latent}. Specifically, for a given value of $Q$, each method ranked users in the test data according to their predicted delinquency probabilities and classified the top-$Q$ users as delinquent. Let $TP$ denote the number of truly delinquent users among the top-$Q$ users, and $n_d$ denote the total number of truly delinquent users. We have
$\text{Precision}=TP/{Q}$,
which measures the fraction of predicted delinquent users who are truly delinquent. $\text{Recall}={TP}/{n_d}$,
which is the fraction of truly delinquent users who are correctly predicted as delinquent. F1 score is the harmonic mean of precision and recall, defined as
$\text{F1}=\frac{2\times\text{Precision}\times\text{Recall}}{\text{Precision}+\text{Recall}}$.
Because there are approximately $300$ delinquent users in the test data, we initially set $Q=300$ in our evaluation. That is, the top-$300$ users with the highest predicted delinquency probabilities were classified as delinquent by each method.
In addition, AUC evaluates the ranking quality of predicted delinquency probabilities across all possible classification thresholds and ranges from 0 to 1, with a larger value indicating better discriminative performance in credit risk prediction. It can be interpreted as the probability that a randomly selected delinquent user receives a higher predicted delinquency probability than a randomly selected non-delinquent user \citep{fang2013predicting}.

\subsection{Benchmark Methods}
\label{sec:eval:benchmark}

We carefully selected representative benchmark methods to evaluate PrivacyCredit from three perspectives. First, we compare PrivacyCredit with an insecure plaintext-learning method to verify whether PrivacyCredit can achieve the same predictive performance as the method that insecurely uses alternative data for credit risk prediction. Second, we compare PrivacyCredit with the common practice adopted by financial institutions, which trains credit risk prediction models using only internally owned credit data, to examine the added predictive value of alternative data. Third, we compare PrivacyCredit with representative privacy-preserving benchmarks to examine whether PrivacyCredit can retain strong predictive performance while ensuring that the learned credit risk prediction model remains centrally stored at the financial institution. We detail the benchmark methods used in our evaluation below.

\textbf{XGBoost-All}: This benchmark applied XGBoost to learn a model from all data in plaintext, i.e., $(X^F,X^A,\bd y)$. XGBoost-All does not satisfy the privacy-preserving constraint because the alternative data $X^A$ are directly transferred to $F$. The performance difference between PrivacyCredit and XGBoost-All indicates whether PrivacyCredit satisfies the lossless constraint.

\textbf{XGBoost-F}: This benchmark applied XGBoost to learn a model from credit data held by $F$ only, i.e., $(X^F,\bd y)$. This benchmark represents the common practice of credit risk prediction at a financial institution. Since the benchmark does not utilize the alternative data held by $A$, the performance difference between PrivacyCredit and XGBoost-F reveals the predictive value of securely utilizing alternative data.

\textbf{SecureBoost}: This benchmark implemented the vertical federated tree boosting method proposed by \citet{cheng2021secureboost}. SecureBoost allows $F$ and $A$ to jointly train a tree boosting model over vertically partitioned data without directly exchanging raw data. However, because the learned model is maintained in a distributed manner, the alternative data provider $A$ knows part of the learned tree structure, especially internal nodes whose split rules are defined on alternative features. Therefore, SecureBoost serves as the main benchmark for evaluating the model-confidentiality advantage of PrivacyCredit.

The following benchmarks are representative PPDP methods. In these benchmarks, the alternative data provider first transformed the alternative data $X^A$ into a privacy-preserving version, denoted by $\tilde{X}^A$, and then shared $\tilde{X}^A$ with the financial institution $F$. The financial institution subsequently trained an XGBoost model using $(X^F,\tilde{X}^A,\bd y)$.\footnote{In standard PPDP applications, explicit identifiers such as user IDs are often removed before data release. However, in our vertically partitioned setting, the financial institution must align each user's traditional data with the corresponding modified alternative data before model training. Therefore, we retained a common user identifier solely for record linkage in our experiments.  This implementation choice enables data alignment, but it also weakens the intended privacy protection of PPDP methods because the released information remains linkable to specific users. }

\textbf{XGBoost-$k$-Anonymity}: This benchmark employed the $k$-anonymity model \citep{sweeney2002k,el2009globally} to modify the alternative data $X^A$ before sharing it with $F$. We treated all alternative mobile features as quasi-identifiers. The alternative data were then generalized to form groups of users who share the same generalized alternative mobile features, with each group containing at least $k$ users. We varied $k$ over three values, $k\in\{3,20,50\}$, where a larger value of $k$ imposes a stronger anonymity requirement and therefore provides stronger privacy protection.

\textbf{XGBoost-$l$-Diversity}: This benchmark employed the $l$-diversity model \citep{machanavajjhala2007diversity,jeon2020proposal} to modify the alternative data $X^A$ before sharing it with $F$. The $l$-diversity model requires each group of users who share the same generalized alternative mobile features to contain sufficiently diverse values of the sensitive attribute. Specifically, we treated the number of phone bill late payments as the sensitive attribute because it directly reflects users' payment behavior and reveals private information related to their creditworthiness. The remaining alternative mobile features were treated as quasi-identifiers. 
We varied $l$ over three values, $l\in\{2,3,4\}$, where a larger value of $l$ imposes a more stringent diversity requirement and therefore provides stronger privacy protection.

\textbf{XGBoost-$t$-Closeness}: This benchmark employed the $t$-closeness model \citep{li2007t,koll2022statistical} to modify the alternative data $X^A$ before sharing it with $F$.  The $t$-closeness model requires the distribution of the sensitive attribute within each group of users who share the same generalized alternative mobile features to be close to its distribution in the overall dataset. As in the $l$-diversity benchmark, we treated the number of phone bill late payments as the sensitive attribute and the remaining alternative mobile features as quasi-identifiers. We varied $t$ over three values, $t\in\{0.1,0.07,0.05\}$, where a smaller value of $t$ imposes a stricter distributional closeness requirement and therefore provides stronger privacy protection.\footnote{For the $l$-diversity and $t$-closeness models, each group of users sharing the same generalized alternative mobile features contains at least 20 users; that is, we set $k=20$.}

\textbf{XGBoost-DP-$\epsilon$}: This benchmark employed a differential privacy-based data publishing method \citep{soria2014enhancing,chen2023global} to modify the alternative data $X^A$ before sharing it with $F$. The privacy loss parameter $\epsilon$ controls the degree of privacy protection, where a smaller value of $\epsilon$ indicates stronger privacy protection but usually leads to greater perturbation of the released data. Because the perturbation introduced by the differential privacy mechanism is randomly generated, we repeated the experiment ten times and reported the average predictive performance across these runs. We varied $\epsilon$ over three values, $\epsilon\in\{10,1,0.1\}$. 

To ensure a fair comparison among these methods, we set the maximum tree depth $D$ to $3$, the learning rate to $0.3$, the number of trees $T$ to $50$, and the regularization parameter $\lambda$ to $1$ for all XGBoost-based methods.
Candidate splits were determined as follows. For each continuous variable (e.g., daily duration of calls), we generated $q$ candidate split thresholds at empirical quantiles of its distribution and set $q=10$ by default. For categorical variables (e.g., type of residence), we converted them into dummy variables and used $0.5$ as the fixed candidate split for each dummy variable. 
We later varied $q$, $D$, and $T$ to examine how candidate split granularity, tree depth, and the number of boosting rounds affect the running time of PrivacyCredit.

\subsection{Predictive Performance and Analysis}
\label{sec:eval:prediction}

Following the evaluation procedure, we conducted experiments to evaluate the predictive performance of each method. Table~\ref{tab:result} reports the results. As shown, PrivacyCredit achieves the same performance as XGBoost-All.
This result empirically verifies the lossless property of PrivacyCredit in practical implementation. It suggests that PrivacyCredit can reproduce the predictive power of a model trained on the insecure plaintext combination of traditional and alternative data, while preserving users' data privacy.

\OneAndAHalfSpacedXI
\begin{table}[t]
\centering
\renewcommand{\arraystretch}{1.25}
\setlength{\tabcolsep}{6pt}
\caption{Predictive Performance Comparison ($Q=300$)}
\label{tab:result}
\resizebox{\columnwidth}{!}{%
\begin{tabular}{L{110pt}L{40pt}C{80pt}C{80pt}C{80pt}C{80pt}}
\hline
\multicolumn{2}{c}{\textbf{Method}}                                                   & \textbf{Precision} & \textbf{Recall} & \textbf{F1 Score} & \textbf{AUC}                 \\ \hline
\multicolumn{2}{c|}{\begin{tabular}[c]{@{}c@{}}\textbf{PrivacyCredit}\\ (our method)\end{tabular}}                                    &    0.4067       &    0.3923    &    0.3993      &          0.8141 \\ \hline
\multicolumn{2}{c|}{XGBoost-All}                                              &    0.4067       &    0.3923    &    0.3993      &          0.8141           \\ \hline
\multicolumn{2}{c|}{XGBoost-F}                                                &    0.3600       &    0.3473    &     0.3535     &           0.7873          \\ \hline
\multicolumn{2}{c|}{SecureBoost}     &    0.4067       &    0.3923    &    0.3993      &          0.8141 \\ \hline
\multicolumn{1}{l|}{\multirow{3}{*}{XGBoost-$k$-Anonymity}} & \multicolumn{1}{l|}{$k=3$}         &    0.3967       &    	0.3826    &     	0.3895     &       0.8061              \\ \cline{2-6} 
\multicolumn{1}{l|}{}                                       & \multicolumn{1}{l|}{$k=20$}         &     0.3933      &    0.3794    &     	0.3863     &         	0.8056            \\ \cline{2-6} 
\multicolumn{1}{l|}{}                                       & \multicolumn{1}{l|}{$k=50$}         &     0.3733      &    0.3601    &     0.3666     &             0.8023        \\ \hline
\multicolumn{1}{l|}{\multirow{3}{*}{XGBoost-$l$-Diversity}} & \multicolumn{1}{l|}{$l=2$}          &    0.3600       & 0.3473 &    0.3535      &     0.7920                \\ \cline{2-6} 
\multicolumn{1}{l|}{}                                       & \multicolumn{1}{l|}{$l=3$}          &     0.3567      &    0.3441    &    0.3502      &            0.7911         \\ \cline{2-6} 
\multicolumn{1}{l|}{}                                       & \multicolumn{1}{l|}{$l=4$}          &     0.3500      &   0.3376     &    0.3437      &   0.7899                  \\ \hline
\multicolumn{1}{l|}{\multirow{3}{*}{XGBoost-$t$-Closeness}} & \multicolumn{1}{l|}{$t=0.1$}        &   	0.3667       &   0.3537    &      0.3601    &   0.7935                 \\ \cline{2-6} 
\multicolumn{1}{l|}{}                                       & \multicolumn{1}{l|}{$t=0.07$}        &    0.3567       &    0.3441    &    0.3502      &           0.7906          \\ \cline{2-6} 
\multicolumn{1}{l|}{}                                       & \multicolumn{1}{l|}{$t=0.05$}        &    0.3533       &     	0.3408  &     0.3470    &    0.7882                 \\ \hline
\multicolumn{1}{l|}{\multirow{3}{*}{XGBoost-DP-$\epsilon$}} & \multicolumn{1}{l|}{$\epsilon=10$} &    0.3953     &     0.3814   &   0.3882     &   0.8029                 \\ \cline{2-6} 
\multicolumn{1}{l|}{}                                       & \multicolumn{1}{l|}{$\epsilon=1$}   &   	0.3703      &    0.3572    &    0.3637      &            0.7937        \\ \cline{2-6} 
\multicolumn{1}{l|}{}                                       & \multicolumn{1}{l|}{$\epsilon=0.1$}  &   	0.3503      &   0.3379     &     0.3440     &      0.7814               \\ \hline      
\end{tabular}%
}
\end{table}
\DoubleSpacedXI

Moreover, PrivacyCredit substantially outperforms XGBoost-F. For example, PrivacyCredit improves recall by 0.045, indicating that among every 100 truly delinquent users, it can correctly identify 4.5 more delinquent users than XGBoost-F. It also increases F1 score and AUC over XGBoost-F by 13.0\% and 3.4\%, respectively. Because XGBoost-F represents the common practice in which a financial institution builds a credit risk prediction model using only its internally owned traditional data, these performance gains demonstrate the predictive value of securely incorporating alternative data through our proposed method.
PrivacyCredit also outperforms the PPDP-based benchmarks. These results suggest that PPDP methods protect privacy at the expense of data utility and consequently degrade the predictive performance of the learned credit risk prediction models. For example, in XGBoost-DP-$\epsilon$, predictive performance improves as $\epsilon$ increases, because a larger $\epsilon$ corresponds to weaker privacy protection and less perturbation to the released data. Nevertheless, even the least stringent setting, XGBoost-DP-$10$, remains inferior to PrivacyCredit. Similar patterns can be observed for the $k$-anonymity, $l$-diversity, and $t$-closeness benchmarks. These findings suggest that directly modifying alternative data before publication inevitably compromises the utility of alternative data for credit risk prediction, thus preventing PPDP methods from satisfying the lossless-performance requirement of the PCRPA problem.

To examine the robustness of our results, we further varied the number of predicted delinquent users $Q$ from $200$ to $400$ in increments of $50$. Table~\ref{tab:rob} reports the results. For readability, among the PPDP-based benchmarks, we report the best-performing setting from each PPDP family, which corresponds to the least stringent privacy protection. The consistent performance between PrivacyCredit and XGBoost-All across different values of $Q$ further confirms the lossless property of PrivacyCredit. Moreover, PrivacyCredit consistently outperforms XGBoost-F and the PPDP-based benchmarks across precision, recall, and F1 score. 

\OneAndAHalfSpacedXI
\begin{table}[t]
\centering
\caption{Predictive Performance Comparison: Varying $Q$}
\label{tab:rob}
\resizebox{\textwidth}{!}{
\begin{tabular}{cc|ccccc|cc}
\hline
Metric & Method & $Q=200$ & $Q=250$ & $Q=300$ & $Q=350$ & $Q=400$ & Mean & Std. \\ \hline
\multirow{8}{*}{Precision}
& \textbf{PrivacyCredit} & 0.4750 & 0.4400 & 0.4067 & 0.3743 & 0.3600 & 0.4112 & 0.0422 \\
& XGBoost-All & 0.4750 & 0.4400 & 0.4067 & 0.3743 & 0.3600 & 0.4112 & 0.0422 \\
& XGBoost-F & 0.4350 & 0.3760  & 0.3600 & 0.3343 & 0.3250 & 0.3661 & 0.0389 \\
& SecureBoost &  0.4750 & 0.4400 & 0.4067 & 0.3743 & 0.3600 & 0.4112 & 0.0422 \\
& XGBoost-$k$-Anonymity ($k=3$) & 0.4700 & 0.4360 &	0.3967  & 0.3714 & 0.3450 & 0.4038 & 0.0447 \\
& XGBoost-$l$-Diversity ($l=2$) & 0.4100 & 0.3840 & 0.3600 & 0.3286 & 0.3200 & 0.3605 & 0.0337 \\
& XGBoost-$t$-Closeness ($t=0.1$)  & 0.4150 & 0.3760 & 0.3667 & 0.3457 & 	0.3300 & 0.3667 & 0.0290 \\
& XGBoost-DP-$\epsilon$ ($\epsilon=10$) & 0.4620 & 0.4232 & 0.3953 & 0.3657 & 	0.3445 & 0.3981 & 	0.0416 \\ \hline

\multirow{8}{*}{Recall}
& \textbf{PrivacyCredit} & 0.3055 & 0.3537 & 0.3923 & 0.4212 & 0.4630  & 0.3871 & 0.0543 \\
& XGBoost-All & 0.3055 & 0.3537 & 0.3923 & 0.4212 & 0.4630  & 0.3871 & 0.0543 \\
& XGBoost-F & 0.2797 & 0.3023 & 0.3473 & 0.3762 & 0.4180 & 0.3447  &  0.0498\\
& SecureBoost & 0.3055 & 0.3537 & 0.3923 & 0.4212 & 0.4630  & 0.3871 & 0.0543 \\
& XGBoost-$k$-Anonymity ($k=3$) & 0.3023 & 0.3505 & 0.3826 & 0.4180 & 0.4437 & 0.3794 & 	0.0499 \\
& XGBoost-$l$-Diversity ($l=2$) & 0.2637 & 0.3087 & 0.3473 & 0.3698 & 0.4116 & 0.3402 & 0.0507 \\
& XGBoost-$t$-Closeness ($t=0.1$) & 0.2669 & 0.3023 & 0.3537 & 0.3891 & 	0.4244 & 0.3473 & 0.0570 \\
& XGBoost-DP-$\epsilon$ ($\epsilon=10$) & 0.2971 & 0.3402 & 0.3814 & 0.4116 & 0.4431  & 0.3747 & 0.0516\\ \hline

\multirow{8}{*}{F1 Score}
& \textbf{PrivacyCredit} & 0.3718 & 0.3922 & 0.3993 & 0.3964 & 0.4051 & 0.3930 & 0.0114 \\
& XGBoost-All & 0.3718 & 0.3922 & 0.3993 & 0.3964 & 0.4051 & 0.3930 & 0.0114 \\
& XGBoost-F & 0.3405 & 0.3351 & 0.3535 & 0.3540 & 0.3657 & 0.3498 & 0.0108 \\
& SecureBoost & 0.3718 & 0.3922 & 0.3993 & 0.3964 & 0.4051 & 0.3930 & 0.0114 \\
& XGBoost-$k$-Anonymity ($k=3$) & 	0.3679 & 0.3886 & 0.3895 & 0.3933 & 0.3882 &  0.3855 & 0.0090 \\
& XGBoost-$l$-Diversity ($l=2$) & 0.3209 & 0.3422 & 0.3535 & 0.3480 & 0.3601  & 0.3449 & 0.0134 \\
& XGBoost-$t$-Closeness ($t=0.1$) & 0.3249 & 0.3351 & 0.3601 & 0.3661 & 0.3713 & 0.3515 & 0.0182 \\
& XGBoost-DP-$\epsilon$ ($\epsilon=10$) & 0.3616 & 0.3772 & 0.3882 & 0.3873 & 0.3876 & 0.3804 & 0.0102\\ \hline
\end{tabular}
}
\end{table}
\DoubleSpacedXI

\subsection{Evaluation of Model Confidentiality}
\label{sec:eval:model_confidentiality}

Financial institutions treat credit risk prediction models as closely guarded trade secrets and therefore require that the learned model remain strictly confidential. Although vertical federated learning methods can securely utilize alternative data without directly exchanging raw data, they train and maintain the learned model in a distributed manner across participating parties. As a result, the alternative data provider observes parts of the learned model. Therefore, in addition to predictive performance, we further evaluate the extent to which the representative vertical federated learning benchmark, SecureBoost, violates the model-confidentiality requirement of the PCRPA problem.

SecureBoost learns an XGBoost model over vertically partitioned data in a distributed manner. An XGBoost model consists of a set of regression trees, and each internal node of a tree contains a split rule, including the split feature and its associated threshold. In SecureBoost, when an internal node is split on an alternative-data feature, the alternative data provider $A$ participates in determining the corresponding split rule, making the split rule of that node observable to $A$. Thus, if $A$ knows the position and split rule of such an internal node, it obtains partial knowledge of the learned model structure. We therefore focus on the exposure of internal nodes in the learned XGBoost model.

To quantify model exposure, we define two metrics: \textit{Node Exposure Count} and \textit{Node Exposure Percentage}. The former counts the number of internal nodes in the learned XGBoost credit risk prediction model that are observable to the alternative data provider $A$, whereas the latter measures the proportion of such exposed nodes among all internal nodes.  Let $\mathcal{B}$ be the set of all internal nodes across the $T$ trees, with $N_{\mathrm{int}}=|\mathcal{B}|$ representing the total number of internal nodes. We further use $\mathcal{E}_{A}$ to denote the set of internal nodes exposed to $A$. The Node Exposure Count, $\mathrm{NEC}_A$, is thus defined as the number of internal nodes in $\mathcal{E}_{A}$, i.e., 
\begin{equation*}
\mathrm{NEC}_A=|\mathcal{E}_{A}|,
\end{equation*}
and the Node Exposure Percentage, $\mathrm{NEP}_A$, is given by
\begin{equation*}
\mathrm{NEP}_A
=
\frac{\mathrm{NEC}_A}{N_{\mathrm{int}}} \times 100\%.
\end{equation*}
A smaller value of $\mathrm{NEC}_A$ or $\mathrm{NEP}_A$ indicates stronger model confidentiality. When $\mathrm{NEC}_A=0$, the alternative data provider has no access to any internal node of the learned model.

We varied the number of trees $T$ and the maximum depth of each tree $D$ to examine model exposure under different XGBoost model settings. Table~\ref{tab:model_confidentiality} reports the comparison between PrivacyCredit and SecureBoost. For PrivacyCredit, $\mathcal{E}_{A}=\emptyset$ across all configurations. This is because the best split at each internal node is obtained and stored only by the financial institution $F$, and the alternative data provider $A$ receives no information about the learned model. Therefore, PrivacyCredit achieves $\mathrm{NEC}_A=0$ and $\mathrm{NEP}_A=0$ regardless of the number of trees $T$ and the maximum tree depth $D$.
In contrast, SecureBoost exposes a large number of internal nodes to $A$, with the $\mathrm{NEP}_A$ ranging from 31.43\% to 51.90\% across the evaluated settings. As shown, the node exposure percentage increases as the number of trees or the maximum tree depth increases. This pattern arises because larger XGBoost models contain more internal nodes, creating more opportunities for alternative-data features to be selected as split variables. In SecureBoost, when such splits are defined on alternative-data features, they become observable to $A$. Therefore, a larger XGBoost model allows $A$ to observe a greater fraction of the learned tree structure. These results empirically support our argument that although SecureBoost can preserve raw data privacy, it cannot guarantee that the learned credit risk prediction model is centrally learned and stored at the financial institution. Thus, SecureBoost does not satisfy the model-confidentiality requirement of the PCRPA problem, whereas PrivacyCredit does.

\OneAndAHalfSpacedXI
\begin{table}[t]
\centering
\caption{Model Confidentiality Under Different XGBoost Model Settings}
\label{tab:model_confidentiality}
\renewcommand{\arraystretch}{1.25}
\setlength{\tabcolsep}{5pt}
\resizebox{\columnwidth}{!}{
\begin{tabular}{ll|cccc|cccc}
\hline
\multirow{2}{*}{Method}
& \multirow{2}{*}{Metric}
& \multicolumn{4}{c|}{Varying Number of Trees $T$ ($D=3$)}
& \multicolumn{4}{c}{Varying Tree Depth $D$ ($T=50$)} \\ \cline{3-10}
& 
& $T=25$ & $T=50$ & $T=75$ & $T=100$
& $D=3$ & $D=4$ & $D=5$ & $D=6$ \\ \hline

\multirow{2}{*}{PrivacyCredit}
& $\mathrm{NEC}_A$ & 0 & 0 & 0 & 0 & 0 & 0 & 0 & 0 \\
& $\mathrm{NEP}_A$ & 0.00\% & 0.00\% & 0.00\% & 0.00\% & 0.00\% & 0.00\% & 0.00\% & 0.00\% \\ \hline

\multirow{2}{*}{SecureBoost}
& $\mathrm{NEC}_A$ & 55 & 122 & 191 & 	266 & 122 & 309 & 700 & 1635 \\
& $\mathrm{NEP}_A$ & 31.43\% & 34.86\% & 36.38\% & 38.00\% & 34.86\% & 41.20\% & 45.16\% & 51.90\% \\ \hline

\end{tabular}
}
\end{table}
\DoubleSpacedXI

\subsection{Computational Efficiency of PrivacyCredit}
\label{sec:eval:efficiency}

Having established the theoretical computational complexity of PrivacyCredit in Section~\ref{subsec:complexity-analysis}, we further evaluate its computational efficiency empirically. The complexity analysis shows that the training time of PrivacyCredit is primarily affected by four key factors: the number of common users $n$, the total number of candidate splits $m$, the number of trees $T$, and the tree depth $D$. In the experiments, we directly varied $n$, $T$, and $D$. For the candidate split set, we used $q$, the number of candidate split thresholds generated for each continuous variable, as a proxy for $m$, since increasing $q$ directly expands the total number of candidate splits.

Following the computational efficiency evaluation procedure commonly used in privacy-preserving computational methods, we varied one parameter at a time while fixing the others at their default values \citep{han2026data}. Table~\ref{tab:eff_default} reports the default parameter setting. Specifically, we set the number of common users to $1{,}000$, the number of candidate split thresholds for each continuous variable to $10$, the number of trees to $50$, and the maximum tree depth to $3$. When evaluating the effect of $n$, we randomly sampled $n$ common users from the training data. When evaluating the effect of $q$, $T$, or $D$, we used the same training sample and varied only the focal parameter. All running time results were measured as the wall-clock training time of PrivacyCredit. We implemented Paillier encryption with a 256-bit modulus. The experiments were conducted on three desktops equipped with Apple M4 chips and 24GB of memory, connected through a wired high-speed local area network.

\OneAndAHalfSpacedXI
\begin{table}[t]
\centering
\caption{Default Parameter Setting for Computational Efficiency Evaluation}
\label{tab:eff_default}
\renewcommand{\arraystretch}{1.2}
\setlength{\tabcolsep}{8pt}
\begin{tabular}{lc}
\hline
\textbf{Parameter} & \textbf{Default Value} \\ \hline
Number of Common Users $n$ & 1,000 \\
Number of Candidate Splits per Continuous Var. $q$ & 10 \\
Number of Trees $T$ & 50 \\ 
Tree Depth $D$ & 3 \\\hline
\end{tabular}
\end{table}
\DoubleSpacedXI

\begin{figure}[b]
\centering
\captionsetup[subfigure]{skip=-4pt}

\begin{subfigure}[t]{0.24\textwidth}
    \centering
    \includegraphics[width=\linewidth]{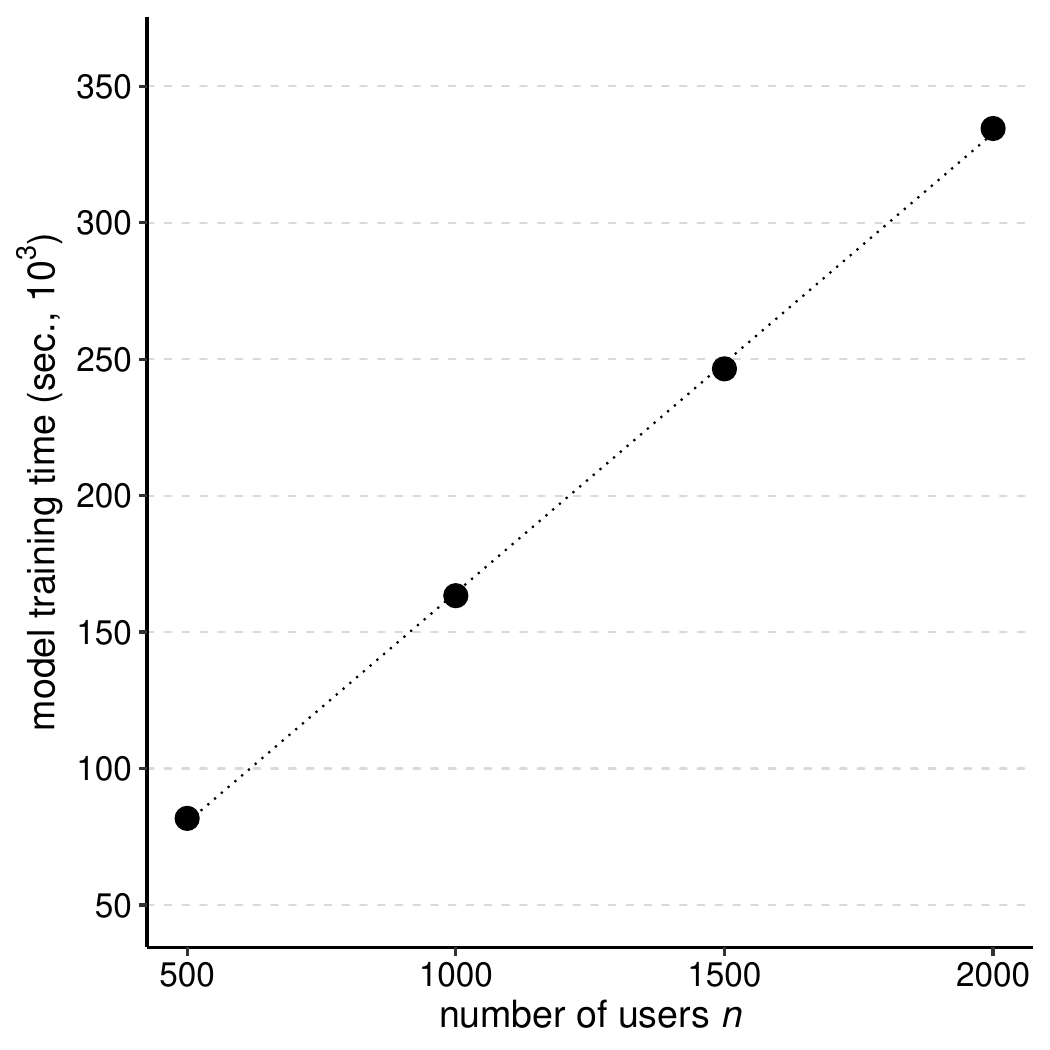}
    \caption{}
    \label{fig:eff_n}
\end{subfigure}
\hfill
\begin{subfigure}[t]{0.24\textwidth}
    \centering
    \includegraphics[width=\linewidth]{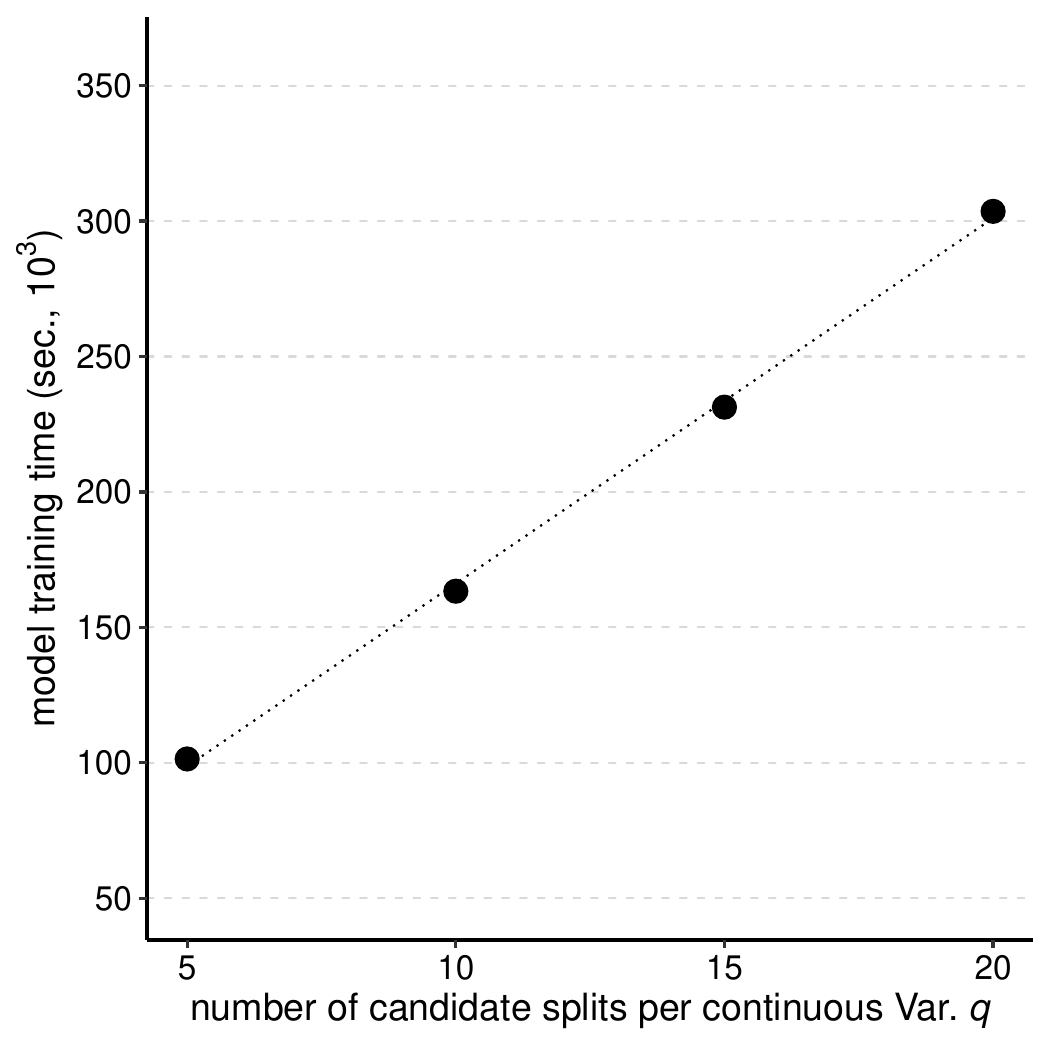}
    \caption{}
    \label{fig:eff_q}
\end{subfigure}
\hfill
\begin{subfigure}[t]{0.24\textwidth}
    \centering
    \includegraphics[width=\linewidth]{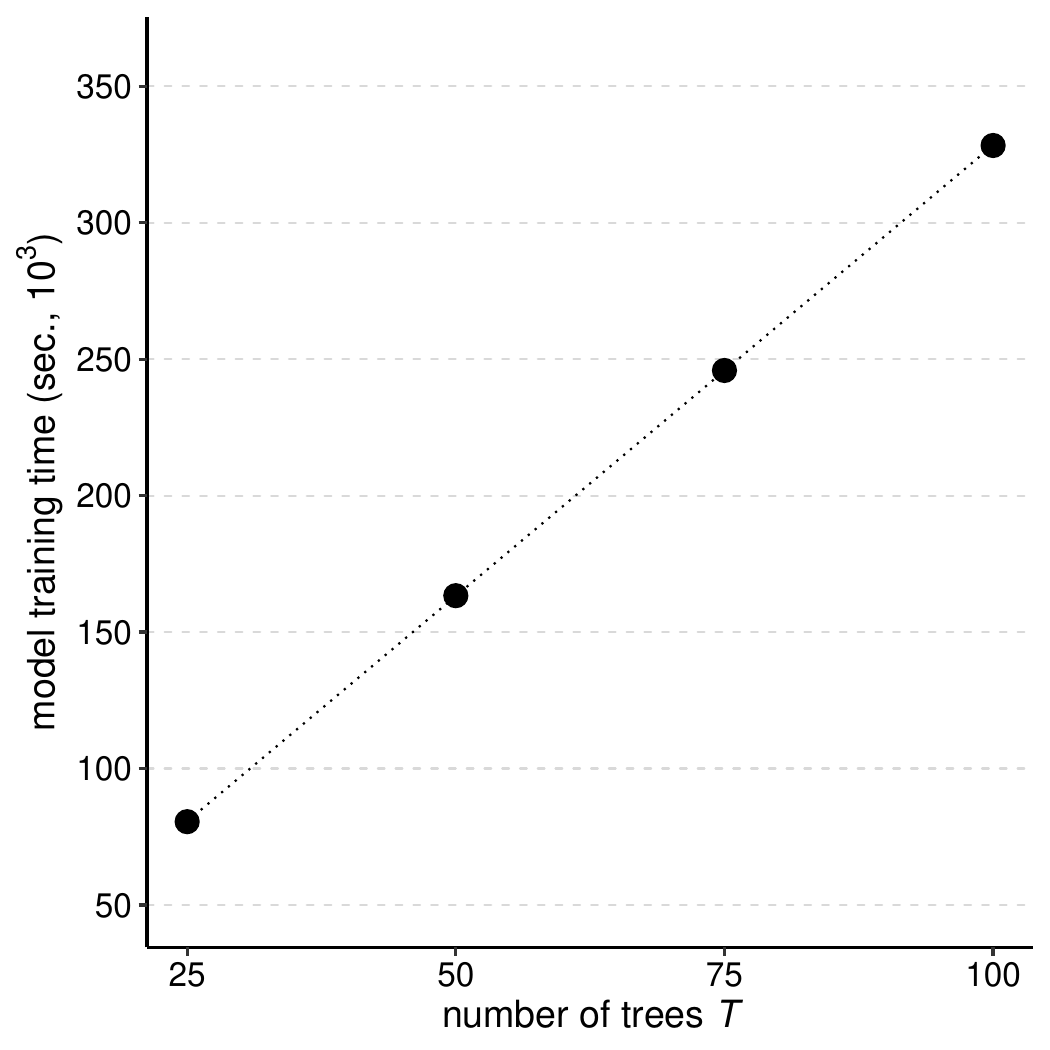}
    \caption{}
    \label{fig:eff_t}
\end{subfigure}
\hfill
\begin{subfigure}[t]{0.24\textwidth}
    \centering
    \includegraphics[width=\linewidth]{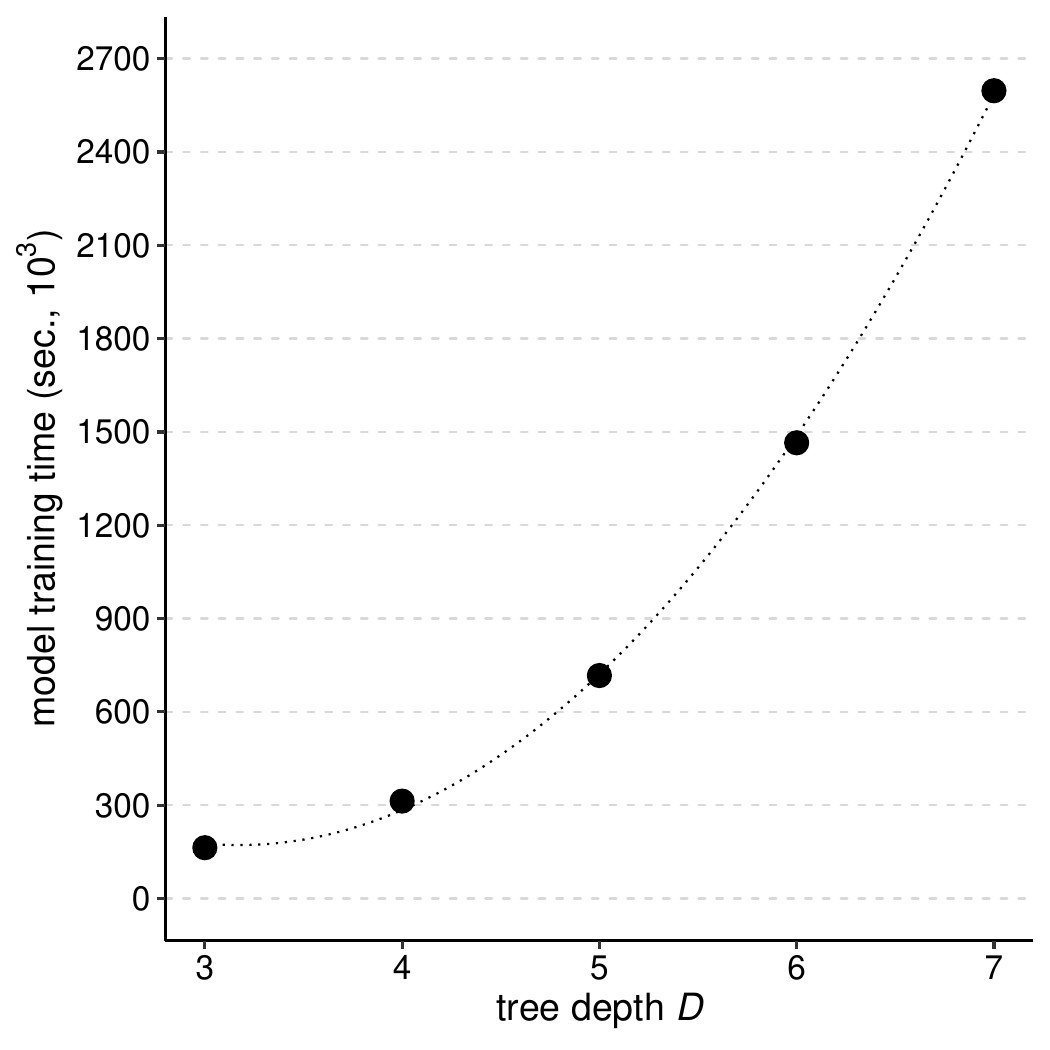}
    \caption{}
    \label{fig:eff_d}
\end{subfigure}

\caption{Computational efficiency of PrivacyCredit under different parameter settings.}
\label{fig:eff}
\end{figure}

Figure~\ref{fig:eff} shows the training time of PrivacyCredit with respect to the four key parameters. Specifically, Figure~\ref{fig:eff_n} exhibits that the training time increases approximately linearly with the number of common users $n$, consistent with the theoretical complexity analysis. Such linear relationship with respect to $n$ suggests that PrivacyCredit can scale predictably when the number of users increases, which is important for real-world credit risk prediction applications involving large user bases.

Figure~\ref{fig:eff_q} reports the effect of candidate split granularity. Recall that we use $q$, the number of candidate split thresholds generated for each continuous variable, as a proxy for the total number of candidate splits $m$. The total candidate split set consists of two parts: candidate splits generated from continuous variables and fixed candidate splits generated from categorical variables after dummy encoding. Increasing $q$ enlarges only the continuous-variable part of the candidate split set, whereas the categorical-variable part remains unchanged. Therefore, the training time increases approximately linearly with $q$, but at a relatively gradual rate because part of the candidate split set is fixed. This pattern is consistent with the complexity analysis, which shows that training time scales linearly with the total number of candidate splits $m$.

Figure~\ref{fig:eff_t} shows the effect of the number of trees $T$. The training time increases approximately linearly with $T$, because increasing $T$ proportionally increases the number of boosting rounds. In each boosting round, PrivacyCredit constructs one encrypted regression tree and updates the model. This result reflects the efficiency-performance tradeoff associated with the number of trees: more boosting rounds give the model greater flexibility and greater capacity to capture predictive patterns, but they also increase the computational burden of PrivacyCredit.

In addition, Figure~\ref{fig:eff_d} shows that the training time increases rapidly as the maximum tree depth $D$ grows. This result is consistent with the theoretical complexity term $2^D$. While deeper trees may capture more complex nonlinear relationships, they also require substantially higher computational costs in a privacy-preserving environment. Therefore, in practical deployment, the efficiency-performance tradeoff of PrivacyCredit can be controlled by choosing a moderate maximum tree depth, which is also common practice in XGBoost-based credit risk prediction.

Collectively, the computational efficiency evaluation empirically complements the theoretical complexity analysis in Section~\ref{subsec:complexity-analysis}. The results demonstrate that PrivacyCredit scales predictably with the number of users, candidate splits, and trees, while the main efficiency bottleneck arises from tree depth. These findings suggest that PrivacyCredit can be practically implemented for privacy-preserving credit risk prediction, especially when standard XGBoost configurations with moderate tree depths are used.

\section{Conclusion}
\label{sec:conclusion}

 \subsection{Summary and Contributions}
\label{sec:conclusion:contribution}

Alternative data have shown significant value in credit risk prediction by enabling lenders to acquire fuller and more accurate profiles of borrowers' creditworthiness. Nevertheless, such data are stored by external entities independent of financial institutions. Directly sharing alternative data with financial institutions for credit risk prediction inevitably infringes on consumer privacy, yet existing credit risk prediction studies largely overlook this issue. To address this limitation, we propose a new problem of privacy-preserving credit risk prediction with alternative data and develop a novel privacy-preserving machine learning method, PrivacyCredit, as a solution to this problem. We theoretically demonstrate the lossless, privacy-preserving, and model-confidential properties of the proposed PrivacyCredit. Through extensive experiments using a real-world credit dataset linked with alternative data, we empirically show that PrivacyCredit achieves the same predictive performance as the insecure plaintext benchmark, outperforms the common practice of credit risk prediction at financial institutions, preserves model confidentiality, and exhibits computational efficiency for practical credit risk prediction.

Our study belongs to the computational genre of design science research in the Information Systems (IS) field, which develops computational algorithms and methods to solve important business and societal problems \citep{rai2017editor,padmanabhan2022editor,fang2025computational}. According to the knowledge contribution framework of design science research \citep{gregor2013positioning}, our study contributes to the extant literature in both the problem and method domains. First, given the sheer size and rapid growth of the consumer credit market, credit risk prediction has attracted increasing attention from IS scholars (e.g., \citealp{wang2021leveraging,yao2025revealed,zhang2025latent}). Our study adds to this stream of research by formulating a new and important problem: Privacy-preserving Credit Risk Prediction with Alternative data (PCRPA). This problem seeks to integrate traditional and alternative data for credit risk prediction while simultaneously protecting consumer privacy, preserving model confidentiality, and maintaining predictive performance without loss. Furthermore, this new problem demands an innovative solution method. Consequently, the second contribution of our study is PrivacyCredit, a novel privacy-preserving machine learning method designed to solve the problem. PrivacyCredit enables a financial institution to learn a credit risk prediction model and perform credit risk inference based on encrypted or randomly masked data, without disclosing users' alternative data to the financial institution or exposing the learned model to the alternative data provider. This capability stems from two key methodological innovations: privacy-preserving best-split finding (Algorithm~\ref{alg:split_finding}) and encrypted leaf routing (Algorithm~\ref{alg:Inf}). The former enables PrivacyCredit to identify the best split for each internal node of an XGBoost tree using encrypted data, allowing the financial institution to construct the credit risk prediction model in a centralized manner. The latter determines the leaf node associated with a new user based on the user's credit data and encrypted alternative data, enabling the financial institution to perform credit risk inference without observing users' alternative data.

\subsection{Implications and Future Work}
\label{sec:conclusion:Implication_FutureWork}

Our study offers several implications for research. First, for the credit risk prediction literature, our study highlights not only the predictive value of alternative data but also the practical challenges associated with using such data. Prior research has primarily emphasized that alternative data can improve credit risk prediction performance \citep{lu2023profit,lee2026benefits}. Our study complements this literature by showing that the use of alternative data also raises important privacy and model-confidentiality concerns. By formulating the PCRPA problem, we provide a structured way to study how alternative data can be used responsibly and effectively in credit risk prediction under the joint requirements of consumer privacy, model confidentiality, and lossless predictive performance.  Second, our proposed method has broader applicability beyond credit risk prediction. Alternative data are increasingly valuable in many business domains, such as employment screening \citep{pallais2014inefficient}, investment decision-making \citep{zhu2019big}, and fraud detection \citep{van2016gotcha}. In these domains, organizations often seek to incorporate external data into proprietary predictive models, creating common requirements for protecting consumer data privacy, learning and storing the model centrally, and preserving predictive performance. 
Therefore, PrivacyCredit provides a foundation for future research on privacy-preserving prediction in a broad range of alternative-data-driven domains.

Our proposed method also offers managerial implications. First, PrivacyCredit provides financial institutions with a practical way to incorporate alternative data into credit risk prediction. By doing so, financial institutions can assess borrowers' credit risk more accurately, thereby better identifying risky borrowers and reducing potential credit losses, while also recognizing creditworthy borrowers and expanding profitable lending opportunities. Given the large scale of the consumer credit market, even modest improvements in credit risk prediction can translate into substantial economic value for financial institutions. In addition, our study creates new opportunities for alternative data providers. Many firms hold alternative data that are informative for credit risk prediction but difficult to share directly because of privacy concerns. PrivacyCredit enables these firms to collaborate with financial institutions without compromising users' data privacy. As a result, alternative data providers can create new revenue streams and business opportunities from their data assets while reducing privacy and reputational risks. Finally, borrowers with limited credit histories can also benefit from our method. By allowing financial institutions to assess borrowers based on a richer information set, PrivacyCredit can improve credit access for creditworthy borrowers who are underserved by traditional credit data. By expanding access to credit for such borrowers, our method can advance financial inclusion and a more equitable consumer credit market, thereby promoting broader social equity and inclusive economic development.

Our study also opens up several directions for future research. First, PrivacyCredit relies on a trusted third party in the Paillier encryption environment to generate key pairs and participate in the protocol. 
Future research could investigate how to eliminate this requirement. One feasible direction is to employ the threshold Paillier cryptosystem, which allows participating parties to collaboratively decrypt ciphertexts without relying on a passive trusted third party. Second, as credit risk prediction models continue to evolve, future research could extend the PCRPA problem to more advanced predictive models beyond XGBoost. For example, generative AI (GenAI) methods create new opportunities for using richer forms of traditional and alternative data, including textual, behavioral, and other unstructured information \citep{feng2023empowering,yin2023finpt}. However, incorporating GenAI into credit risk prediction also raises new challenges. Future research could examine how to enable financial institutions to use both traditional and alternative data in GenAI-based prediction without disclosing users' private information, what types of model information should be treated as confidential in the GenAI context, and how such confidentiality can be preserved. Addressing these questions would extend the idea of PrivacyCredit to a new generation of credit risk prediction models.

\bibliographystyle{informs2014} 
\bibliography{references} 


\ECSwitch


\ECHead{Supplementary Materials}
%

%
%
%
\begin{APPENDICES}

\section{Proof of Lemmas}
\label{apd:lma}

\subsection{Proof of Lemma~\ref{lma:score_s}}
\label{apd:lma:score}

\begin{align*}
&\quad [[\frac{(N_1)^2}{M_1}]]^{\beta_1}\times [[-\frac{N_1}{M_1}]]^{2\alpha_1\beta_1}\times[[\frac{1}{M_1}]]^{\alpha_1^2\beta_1}\times[[\frac{(N_2)^2}{M_2}]]^{\beta_2}\times[[-\frac{N_2}{M_2}]]^{2\alpha_2\beta_2}\times[[\frac{1}{M_2}]]^{ \alpha_2^2\beta_2}\\
\overset{\quad\quad}{=}&\quad [[\frac{(G^L_S+\alpha_1)^2}{\beta_1(H^L_S+\lambda)}]]^{\beta_1}\times [[-\frac{G^L_S+\alpha_1}{\beta_1(H^L_S+\lambda)}]]^{2\alpha_1\beta_1}\times[[\frac{1}{\beta_1(H^L_S+\lambda)}]]^{\alpha_1^2\beta_1}\\
&\times[[\frac{(G^R_S+\alpha_2)^2}{\beta_2(H^R_S+\lambda)}]]^{\beta_2}\times[[-\frac{G^R_S+\alpha_2}{\beta_2(H^R_S+\lambda)}]]^{2\alpha_2\beta_2}\times[[\frac{1}{\beta_2(H^R_S+\lambda)}]]^{ \alpha_2^2\beta_2}\\
\overset{Eq.\eqref{eq:pai_mul}}{=}&\quad [[\frac{\beta_1(G^L_S+\alpha_1)^2}{\beta_1(H^L_S+\lambda)}]]\times[[-\frac{2\alpha_1\beta_1(G^L_S+\alpha_1)}{\beta_1(H^L_S+\lambda)}]]\times[[\frac{\alpha_1^2\beta_1}{\beta_1(H^L_S+\lambda)}]]\\
&\times[[\frac{\beta_2(G^R_S+\alpha_2)^2}{\beta_2(H^R_S+\lambda)}]]\times[[-\frac{2\alpha_2\beta_2(G^R_S+\alpha_2)}{\beta_2(H^R_S+\lambda)}]]\times[[\frac{\alpha_2^2\beta_2}{\beta_2(H^R_S+\lambda)}]]\\
\overset{Eq.\eqref{eq:pai_sub}}{=}&\quad [[\frac{(G^L_S+\alpha_1)^2-2\alpha_1(G^L_S+\alpha_1)}{H^L_S+\lambda}]]\times[[\frac{\alpha_1^2}{H^L_S+\lambda}]]\times[[\frac{(G^R_S+\alpha_2)^2-2\alpha_2(G^R_S+\alpha_2)}{H^R_S+\lambda}]]\times[[\frac{\alpha_2^2}{H^R_S+\lambda}]]\\
\overset{\quad\quad}{=}&\quad [[\frac{(G^L_S)^2-\alpha_1^2}{H^L_S+\lambda}]]\times[[\frac{\alpha_1^2}{H^L_S+\lambda}]]\times[[\frac{(G^R_S)^2-\alpha_2^2}{H^R_S+\lambda}]]\times[[\frac{\alpha_2^2}{H^R_S+\lambda}]]\\
\overset{Eq.\eqref{eq:pai_add}}{=}&\quad [[\frac{(G^L_S)^2-\alpha_1^2+\alpha_1^2}{H^L_S+\lambda}]]
\times[[\frac{(G^R_S)^2-\alpha_2^2+\alpha_2^2}{H^R_S+\lambda}]]\quad
\overset{Eq.\eqref{eq:pai_add}}{=}\quad [[\frac{(G^L_S)^2}{H^L_S+\lambda}
+\frac{(G^R_S)^2}{H^R_S+\lambda}]]\quad
\overset{Eq.\eqref{eq:score_s}}{=} \quad[[score_S]]
\end{align*}

\subsection{Proof of Lemma~\ref{lma:weight}}
\label{apd:lma:weight}

\begin{align*}
&\quad		[[-\frac{N_k}{M_k}]]^{\beta_k}\times[[\frac{1}{M_k}]]^{\alpha_k\beta_k}	\overset{\quad\quad}{=}\quad	[[-\frac{G^k+\alpha_k}{\beta_k(H^k+\lambda)}]]^{\beta_k}\times[[\frac{1}{\beta_k(H^k+\lambda)}]]^{\alpha_k\beta_k}\\
\overset{Eq.\eqref{eq:pai_mul}}{=}&\quad[[\frac{-\beta_k(G^k+\alpha_k)}{\beta_k(H^k+\lambda)}]]\times[[\frac{\alpha_k\beta_k}{\beta_k(H^k+\lambda)}]]\quad
\overset{Eq.\eqref{eq:pai_add}}{=}\quad[[\frac{-(G^k+\alpha_k)}{H^k+\lambda}+\frac{\alpha_k}{H^k+\lambda}]]\quad
\overset{\quad\quad}{=}\quad[[\frac{-G^k}{H^k+\lambda}]]\quad
\overset{Eq.\eqref{eq:fitted_value}}{=} \quad[[\omega_k]]
\end{align*}

\subsection{Proof of Lemma~\ref{lma:update}}
\label{apd:lma:update}

\begin{align*}
&\quad\prod_k([[W_k\times O_k]]\times[[-I_{i,t}^k]]^{\alpha_k}\times[[-\omega_{t,k}]]^{\beta_k}\times[[-\alpha_k\beta_k]])\\
\overset{\quad\quad}{=}&\quad\prod_k([[(\omega_{t,k}+\alpha_k)(I_{i,t}^k+\beta_k)]]\times[[-I_{i,t}^k]]^{\alpha_k}\times[[-\omega_{t,k}]]^{\beta_k}\times[[-\alpha_k\beta_k]])\\
\overset{Eq.\eqref{eq:pai_mul}}{=}&\quad\prod_k([[(\omega_{t,k}+\alpha_k)(I_{i,t}^k+\beta_k)]]\times[[-\alpha_kI_{i,t}^k]]\times[[-\beta_k\omega_{t,k}]]\times[[-\alpha_k\beta_k]])\\
\overset{Eq.\eqref{eq:pai_sub}}{=}&\quad\prod_k([[(\omega_{t,k}+\alpha_k)(I_{i,t}^k+\beta_k)-\alpha_kI_{i,t}^k-\beta_k\omega_{t,k}-\alpha_k\beta_k]])
\overset{Eq.\eqref{eq:pai_add}}{=}\quad[[\sum_k\omega_{t,k}I_{i,t}^k]]\quad
\overset{Eq.\eqref{eq:xgboost_sum}}{=} \quad[[\omega_{t,q_t(\x_i)}]]
\end{align*}

\section{Proof of Proposition~\ref{thm:pcrpa_constraints}}
\label{apd:thm}

\proof{Proof.}
We prove the three properties in Proposition~\ref{thm:pcrpa_constraints} in turn.

\textit{Losslessness.} PrivacyCredit implements the same learning and inference operations as vanilla XGBoost, but performs the required computations over ciphertexts. We first consider model learning. For any candidate split $S$, the encrypted comparison protocol correctly produces encrypted indicators whose plaintexts are identical to the corresponding plaintext split indicators. Therefore, the plaintexts underlying $[[I^L_S]]$ and $[[I^R_S]]$ are exactly the left and right instance sets that would be obtained by applying split $S$ in vanilla XGBoost. Given these indicators, the homomorphic operations in Algorithm~\ref{alg:split_finding} compute encrypted sufficient statistics and encrypted split scores whose plaintexts coincide with their plaintext counterparts in vanilla XGBoost. Hence, applying the encrypted \textit{argmax} protocol to these encrypted scores selects the same best split as vanilla XGBoost.

The same argument applies to leaf-weight computation. The homomorphic operations and Lemma~\ref{lma:weight} ensure that the plaintext value underlying each encrypted leaf weight $[[\omega_{t,k}]]$ is exactly the leaf weight that vanilla XGBoost would compute. Moreover, Algorithm~\ref{alg:update} updates each $p_i^{(t)}$, and hence each pair of first- and second-order derivatives, using the same leaf contribution as vanilla XGBoost. Therefore, by induction over tree nodes and boosting iterations, PrivacyCredit learns the same tree structures and the same underlying leaf weights as the model $\M^\circ$ learned from the insecure plaintext combination of $(X^F,\bd y)$ and $X^A$.

During inference, Algorithm~\ref{alg:Inf} uses encrypted comparisons to route a new borrower to the same leaf in each tree as vanilla XGBoost. The homomorphic additions then aggregate the same leaf contributions across trees. Thus, for any input instance, PrivacyCredit produces the same log-odds and the same predicted probability as $\M^\circ$. Consequently, the model $\M$ learned by PrivacyCredit achieves identical predictive performance to $\M^\circ$.

\textit{Privacy preservation.} We next show that PrivacyCredit does not disclose traditional data held by $F$ or alternative data held by $A$. In the regression tree learning stage, for each candidate split $S\in\s^A$, $A$ sends only the ciphertext $[[I^<_S]]$ to $F$. Because $F$ cannot decrypt this ciphertext, it learns no plaintext information about users' alternative data. The subsequent construction of $[[I^L_S]]$ and $[[I^R_S]]$ relies only on homomorphic operations and the encrypted comparison protocol, whose outputs remain encrypted. Hence, these operations do not reveal whether any particular user falls into the left or right branch of a split.

The quantities $g$ and $h$, which depend on users' credit outcomes held by $F$, are never disclosed in plaintext. To compute split scores and leaf weights, $F$ sends to $C$ only randomly masked encrypted quantities, such as $[[G^L_S+\alpha_1]]$, $[[G^R_S+\alpha_2]]$, $[[\beta_1(H^L_S+\lambda)]]$, $[[\beta_2(H^R_S+\lambda)]]$, $[[G^k+\alpha_k]]$, and $[[\beta_k(H^k+\lambda)]]$. Because the random masks are generated by $F$ and unknown to $C$, the decrypted values observed by $C$ do not reveal the underlying sufficient statistics. In addition, the encrypted \textit{argmax} protocol reveals only the best split to $F$, and the best split itself does not disclose any individual user's private data.

In the parameter updating stage, Algorithm~\ref{alg:update} similarly sends only masked quantities to $C$, including $[[\omega_{t,k}+\alpha_k]]$, $[[I^k_{i,t}+\beta_k]]$, and $[[\omega_{t,q_t(\x_i)}+\xi]]$. The random masks prevent $C$ from recovering the corresponding unmasked values. Although $F$ obtains the plaintext leaf contribution $\omega_{t,q_t(\x_i)}$, it cannot infer the sensitive leaf membership indicators because the individual leaf weights and membership indicators are not revealed in plaintext.

Finally, in the inference stage, $A$ sends only encrypted alternative features $[[\x^A_{\mathrm{new}}]]$ to $F$. The branch indicators $[[\delta_b]]$ and leaf indicators $[[\delta^k_t]]$ are computed through encrypted comparisons and homomorphic operations, and thus reveal no plaintext alternative features to $F$. When obtaining the final log-odds, $F$ sends only the masked ciphertext $[[\theta_{\mathrm{new}}+\rho]]$ to $C$, where $\rho$ is known only to $F$. Therefore, $C$ cannot infer $\theta_{\mathrm{new}}$. Taken together, throughout model learning and inference, all information received by a party other than the data owner is either encrypted or randomly masked. Thus, no traditional data $(X^F,\bd y)$ held by $F$ are leaked to $A$, and no alternative data $X^A$ held by $A$ are leaked to $F$.

\textit{Model confidentiality.} An XGBoost model is fully determined by the split selected at each internal node and the weight assigned to each leaf. In PrivacyCredit, the best split at each internal node is obtained through the encrypted \textit{argmax} protocol, which reveals the selected split only to $F$. Other parties observe only ciphertexts or randomly masked intermediate quantities and therefore learn no information about the tree structure.

For leaf weights, $F$ computes only encrypted weights using Lemma~\ref{lma:weight}. During this computation, $C$ receives only randomly masked aggregate statistics, such as $[[G^k+\alpha_k]]$ and $[[\beta_k(H^k+\lambda)]]$, and hence cannot recover the unmasked statistics or the resulting leaf weights. The alternative data owner $A$ receives no information about the learned split scores, tree structure, or leaf weights. Therefore, no party other than $F$ obtains information about the learned model $\M$.

Since the model structure and leaf weights together fully characterize $\M$, and neither component is disclosed to any party other than $F$, PrivacyCredit satisfies the model-confidential constraint. Combining the three parts above, PrivacyCredit satisfies the lossless, privacy-preserving, and model-confidential constraints of the PCRPA problem.
\clearpage

\section{Detailed Computational Complexity Analysis}
\label{app:complexity}

This appendix provides the detailed computational complexity analysis of PrivacyCredit. We analyze each component of the training and inference procedures and then derive the total complexity.

\subsection{Notation}

Let $n$ denote the number of common users in the training set. Let $d_F$ and $d_A$ denote the numbers of traditional features held by the financial institution $F$ and alternative features held by the alternative data owner $A$, respectively. Let
$$
\mathcal{S}_F,\quad \mathcal{S}_A,\quad \mathcal{S}=\mathcal{S}_F\cup \mathcal{S}_A
$$
denote the candidate split sets generated from traditional features, alternative features, and all features. We define
$$
m_F = |\mathcal{S}_F|,\quad 
m_A = |\mathcal{S}_A|,\quad 
m = |\mathcal{S}| = m_F + m_A.
$$
If $q$ candidate thresholds are generated per feature, then
$$
m = O(q(d_F+d_A)).
$$

Let $T$ be the number of boosting trees and $D$ be the maximum depth of each tree. A full binary tree of depth $D$ has at most
$$
B_D = 2^D - 1
$$
internal nodes and
$$
K_D = 2^D
$$
leaf nodes. If the learned tree is not full, $B_D$ and $K_D$ can be replaced by the actual numbers of internal and leaf nodes.

We use the following primitive cost notation.

\begin{table}[htbp]
\centering
\caption{Notation for primitive cryptographic costs}
\label{tab:primitive-costs}
\begin{tabular}{p{0.28\linewidth}p{0.62\linewidth}}
\hline
\textbf{Symbol} & \textbf{Meaning} \\
\hline
$C_{\mathrm{Enc}}$ 
& Cost of one Paillier encryption \\

$C_{\mathrm{Dec}}$ 
& Cost of one Paillier decryption \\

$C_{\mathrm{Add}}$ 
& Cost of one homomorphic addition or subtraction \\

$C_{\mathrm{SMul}}$ 
& Cost of one homomorphic scalar multiplication \\

$C_{\mathrm{Dot}}(n)$ 
& Cost of one encrypted dot product between a plaintext vector and an encrypted vector of length $n$ \\

$C_{\mathrm{Cmp}}(\ell)$ 
& Cost of one Veugen encrypted comparison on $\ell$-bit integers \\

$C_{\mathrm{Arg}}(m,\ell_s)$ 
& Cost of one Bost encrypted argmax over $m$ encrypted values of bit length $\ell_s$ \\
\hline
\end{tabular}
\end{table}

A homomorphic dot product between a plaintext vector and an encrypted vector of length $n$ requires $n$ scalar multiplications and $n-1$ homomorphic additions. Therefore, we write
$$
C_{\mathrm{Dot}}(n)
=
O\big(n(C_{\mathrm{SMul}}+C_{\mathrm{Add}})\big).
$$

Veugen's comparison protocol is linear in the bit length of the compared integers. Thus,
$$
C_{\mathrm{Cmp}}(\ell)=O(\ell)
$$
up to cryptographic constants. Bost's encrypted argmax performs $m-1$ encrypted comparisons and a linear number of masking and refresh operations. Hence,
$$
C_{\mathrm{Arg}}(m,\ell_s)
=
O\big((m-1)C_{\mathrm{Cmp}}(\ell_s)+m\big).
$$

\subsection{Algorithm 1: Obtaining Encrypted Split Indicators}

Algorithm 1 allows $F$ to obtain encrypted split-indicator vectors
$$
[[I_S^{<}]],\quad \forall S\in\mathcal{S}.
$$
For each candidate split $S$, the indicator vector $I_S^{<}\in\{0,1\}^n$ has length $n$. Constructing this vector requires $n$ plaintext comparisons between feature values and the split threshold. Encrypting the vector requires $n$ Paillier encryptions.

Therefore, Algorithm 1 requires
$$
O(nm)
$$
plaintext comparisons and
$$
nm C_{\mathrm{Enc}}
$$
Paillier encryptions. Its total computational cost is
$$
C_1
=
O(nm) + nm C_{\mathrm{Enc}}.
$$

The communication cost comes from $A$ sending encrypted alternative-data split indicators to $F$. Since $A$ owns $m_A$ candidate splits, this communication cost is
$$
O(nm_A)
$$
Paillier ciphertexts.

\subsection{Algorithm 2: Privacy-Preserving Best Split Selection}

Algorithm 2 is invoked once at each internal node. Given the current encrypted node-membership vector $[[I]]$, it evaluates all candidate splits and selects the split with the maximum encrypted score.

\subsubsection{Per-Split Cost.}

For one candidate split $S$, Algorithm 2 first constructs encrypted left and right child membership indicators,
$$
[[I_S^L]],\quad [[I_S^R]].
$$
For each user $i$,
$$
I_{i,S}^{L}=1
\quad\text{if and only if}\quad
I_i+I_{i,S}^{<}=2.
$$
Since both $I_i$ and $I_{i,S}^{<}$ are binary, the value $I_i+I_{i,S}^{<}$ lies in $\{0,1,2\}$. Therefore, this equality check can be implemented using a constant-bit encrypted comparison. Constructing $[[I_S^L]]$ and $[[I_S^R]]$ costs
$$
n C_{\mathrm{Cmp}}(2) + O(n).
$$

Next, Algorithm 2 computes four encrypted sufficient statistics:
$$
[[G_S^L]],\quad [[G_S^R]],\quad [[H_S^L]],\quad [[H_S^R]].
$$
Each is an encrypted dot product between a plaintext gradient or Hessian vector and an encrypted indicator vector. Thus, this step costs
$$
4C_{\mathrm{Dot}}(n).
$$

Finally, Algorithm 2 computes the encrypted split score
$$
[[\mathrm{score}_S]]
=
\left[
\left[
\frac{(G_S^L)^2}{H_S^L+\lambda}
+
\frac{(G_S^R)^2}{H_S^R+\lambda}
\right]
\right].
$$
Because Paillier encryption does not support division, PrivacyCredit uses random masking and the trusted authority $C$ to compute the required encrypted division terms. This step involves a constant number of encryptions, decryptions, and homomorphic operations. Denote this constant cost by $C_{\mathrm{Score}}$.

Therefore, the per-split cost is
$$
C_{\mathrm{split}}
=
nC_{\mathrm{Cmp}}(2)
+
4C_{\mathrm{Dot}}(n)
+
O(n)
+
C_{\mathrm{Score}}.
$$
Since $C_{\mathrm{Cmp}}(2)$ is constant and $C_{\mathrm{Dot}}(n)=O(n)$, we have
$$
C_{\mathrm{split}}=O(n).
$$

\subsubsection{Per-Node Cost.}

At one internal node, Algorithm 2 evaluates all $m$ candidate splits. The encrypted score computation therefore costs
$$
m C_{\mathrm{split}}.
$$
After all encrypted split scores are obtained, PrivacyCredit applies Bost's encrypted argmax protocol. This costs
$$
C_{\mathrm{Arg}}(m,\ell_s)
=
O\big((m-1)C_{\mathrm{Cmp}}(\ell_s)+m\big),
$$
where $\ell_s$ is the bit length of the fixed-point encoded split scores.

Thus, the total cost of Algorithm 2 at one internal node is
$$
C_2^{\mathrm{node}}
=
mC_{\mathrm{split}}
+
C_{\mathrm{Arg}}(m,\ell_s).
$$
Substituting the per-split cost gives
$$
C_2^{\mathrm{node}}
=
m\left[
nC_{\mathrm{Cmp}}(2)
+
4C_{\mathrm{Dot}}(n)
+
O(n)
+
C_{\mathrm{Score}}
\right]
+
C_{\mathrm{Arg}}(m,\ell_s).
$$
Asymptotically, this simplifies to
$$
C_2^{\mathrm{node}}
=
O(mn + m\ell_s).
$$

\subsection{Recursive Tree Construction and Leaf Weight Computation}

A full binary tree of depth $D$ contains at most $2^D-1$ internal nodes. Since Algorithm 2 is invoked at every internal node, the total split-selection cost for one tree is
$$
(2^D-1)C_2^{\mathrm{node}}.
$$

After the structure of the tree is determined, PrivacyCredit computes encrypted weights for the leaf nodes. For leaf $k$, the XGBoost leaf weight is
$$
\omega_k
=
-
\frac{G_k}{H_k+\lambda}.
$$
Computing $[[G_k]]$ and $[[H_k]]$ requires two encrypted dot products of length $n$. The masked division procedure requires only a constant number of additional cryptographic operations. Denote this constant cost by $C_{\mathrm{LeafDiv}}$. The cost per leaf is
$$
C_{\mathrm{leaf}}
=
2C_{\mathrm{Dot}}(n)
+
C_{\mathrm{LeafDiv}}.
$$

A full tree has at most $2^D$ leaves. Hence, the total leaf-weight computation cost is
$$
2^D C_{\mathrm{leaf}}.
$$
Therefore, the cost of constructing one tree is
$$
C_{\mathrm{tree}}
=
(2^D-1)C_2^{\mathrm{node}}
+
2^D C_{\mathrm{leaf}}.
$$
Using
$$
C_2^{\mathrm{node}}=O(mn+m\ell_s)
$$
and
$$
C_{\mathrm{leaf}}=O(n),
$$
we obtain
$$
C_{\mathrm{tree}}
=
O\left(
2^D[m(n+\ell_s)+n]
\right).
$$
When $\ell_s$ is fixed and $m$ is moderately large, this becomes
$$
C_{\mathrm{tree}}=O(2^Dmn).
$$

\subsection{Algorithm 3: Privacy-Preserving Parameter Updating}

After constructing the $t$-th tree, PrivacyCredit updates each user's prediction and derivatives. For user $i$, Algorithm 3 computes
$$
[[\omega_{t,q_t(x_i)}]]
=
\left[
\left[
\sum_{k=1}^{K_t}
\omega_{t,k} I_{i,t}^{k}
\right]
\right],
$$
where $K_t$ is the number of leaves in tree $t$. Since both $[[ \omega_{t,k} ]]$ and $[[ I_{i,t}^{k} ]]$ are encrypted, PrivacyCredit uses the masking protocol in Lemma 3. For each pair $(i,k)$, this requires a constant number of encrypted operations and masked decryptions. Let this constant cost be $C_{\mathrm{Prod}}$.

For one user, the cost is
$$
K_t C_{\mathrm{Prod}}.
$$
For all $n$ users, the cost is
$$
C_3^{(t)}
=
nK_t C_{\mathrm{Prod}} + O(n).
$$
Since $K_t\leq 2^D$, we have
$$
C_3^{(t)}=O(n2^D).
$$
This step is linear in the number of users and exponential in tree depth, but it does not depend on the number of candidate splits $m$.

\subsection{Total Training Complexity}

Training consists of Algorithm 1 followed by $T$ boosting rounds. Each boosting round constructs one tree and then updates the predictions and derivatives.

The total training cost is
$$
C_{\mathrm{train}}
=
C_1
+
\sum_{t=1}^{T}
\left[
C_{\mathrm{tree}}^{(t)}
+
C_3^{(t)}
\right].
$$
Using the full-tree upper bound, we have
$$
C_{\mathrm{train}}
=
O(nm)
+
O\left(
T2^D[m(n+\ell_s)+n]
\right).
$$
When $\ell_s$ is fixed and $m$ is not trivially small, the dominant term is
$$
C_{\mathrm{train}}
=
O(T2^Dmn).
$$
Because $m=O(q(d_F+d_A))$, this can also be written as
$$
C_{\mathrm{train}}
=
O\left(
T2^D q(d_F+d_A)n
\right).
$$
Thus, training complexity is linear in the number of users $n$, linear in the number of candidate splits $m$, linear in the number of trees $T$, and exponential in the tree depth $D$.

\subsection{Algorithm 4: Privacy-Preserving Inference}

We next analyze inference for one new borrower. Let $B_t$ and $K_t$ denote the numbers of internal and leaf nodes in tree $t$. At an internal node whose split feature belongs to $F$, the comparison can be done locally by $F$, followed by encryption of the branch indicator. At an internal node whose split feature belongs to $A$, PrivacyCredit performs an encrypted comparison between the encrypted alternative feature and the split threshold. If the encoded feature has bit length $\ell_x$, this costs
$$
C_{\mathrm{Cmp}}(\ell_x).
$$

After branch indicators are computed, PrivacyCredit computes leaf indicators. For each leaf, it sums the encrypted branch indicators along the path and compares the encrypted path sum with the path length. Since the path length is at most $D$, the comparison bit length is
$$
\ell_p = \lceil \log_2(D+1)\rceil.
$$
Thus, computing all leaf indicators for tree $t$ costs
$$
O\big(K_t C_{\mathrm{Cmp}}(\ell_p) + K_tD\big).
$$

Finally, PrivacyCredit combines encrypted leaf indicators with encrypted leaf weights using the Lemma 3 masking protocol. This requires $O(K_t)$ constant-cost encrypted products.

Therefore, the inference cost for tree $t$ is
$$
C_{\mathrm{infer}}^{(t)}
=
O\left(
B_t C_{\mathrm{Cmp}}(\ell_x)
+
K_t C_{\mathrm{Cmp}}(\ell_p)
+
K_tD
+
K_t
\right).
$$
Using the full-tree upper bounds $B_t=O(2^D)$ and $K_t=O(2^D)$, we obtain
$$
C_{\mathrm{infer}}^{(t)}
=
O\left(
2^D[
C_{\mathrm{Cmp}}(\ell_x)
+
C_{\mathrm{Cmp}}(\ell_p)
+
D
]
\right).
$$
Across $T$ trees, the inference cost for one borrower is
$$
C_{\mathrm{infer}}
=
O\left(
T2^D[
C_{\mathrm{Cmp}}(\ell_x)
+
C_{\mathrm{Cmp}}(\ell_p)
+
D
]
\right).
$$
When the comparison bit lengths are fixed and $D$ is small, this simplifies to
$$
C_{\mathrm{infer}}=O(T2^D).
$$
For $n_{\mathrm{test}}$ test borrowers, the total inference cost is
$$
O(n_{\mathrm{test}}T2^D).
$$

Unlike training, inference complexity does not depend on the training-set size $n$ or the number of candidate splits $m$. Once the model is trained, prediction depends only on the number of trees, tree depth, and encrypted comparison bit lengths.

\subsection{Summary}

Table~\ref{tab:complexity-summary} summarizes the main complexity results.

\begin{table}[htbp]
\centering
\caption{Summary of computational complexity}
\label{tab:complexity-summary}
\begin{tabular}{p{0.42\linewidth}p{0.48\linewidth}}
\hline
\textbf{Procedure} & \textbf{Complexity} \\
\hline
Algorithm 1 
& $O(nm)$ encryptions \\

Algorithm 2 at one node 
& $O(mn+m\ell_s)$ \\

One tree construction 
& $O(2^D[m(n+\ell_s)+n])$ \\

Algorithm 3 per tree 
& $O(n2^D)$ \\

Total training 
& $O(nm)+O(T2^D[m(n+\ell_s)+n])$ \\

Simplified training 
& $O(T2^Dmn)$ \\

Inference for one borrower 
& $O(T2^D)$, fixed bit lengths \\

Inference for $n_{\mathrm{test}}$ borrowers 
& $O(n_{\mathrm{test}}T2^D)$ \\
\hline
\end{tabular}
\end{table}

The results imply that training time is primarily driven by four factors: the number of training users, the number of candidate splits, the number of trees, and tree depth. In contrast, inference time is independent of the training-set size and candidate split set after the model has been learned.

\end{APPENDICES}

\ACKNOWLEDGMENT{}






%
%
%

\end{document}